%% file: main.tex
\documentclass[11pt]{article}
\PassOptionsToPackage{table}{xcolor} 
\usepackage[round,authoryear]{natbib}
\PassOptionsToPackage{no-math}{fontspec}
\usepackage{metacircle}
\usepackage{float}
\usepackage{capt-of}
\usepackage{listings}
\usepackage{needspace}

\graphicspath{{figures/}}

\definecolor{kbpaper}{HTML}{FBFAF5}
\definecolor{kbrule}{HTML}{3A3A3A}
\definecolor{kbink}{HTML}{1F1F1F}
\definecolor{kbobs}{HTML}{D55E00}
\definecolor{kbqnt}{HTML}{0072B2}
\definecolor{kbrnk}{HTML}{009E73}
\definecolor{kbaccopt}{HTML}{56B4E9}
\definecolor{kbaccdat}{HTML}{D9B300}
\definecolor{kbaccarc}{HTML}{CC79A7}
\definecolor{kbsilver}{HTML}{DBDFE3}
\tikzset{kbchip/.style={rounded corners=1.6pt, inner xsep=3pt, inner ysep=1.1pt,
  font=\fontsize{7.2}{8}\selectfont\sffamily}}
\newcommand{\kbob}[1]{\textcolor{kbobs}{#1}}
\newcommand{\kbqt}[1]{\textcolor{kbqnt}{#1}}
\newcommand{\kbrk}[1]{\par\vspace{2.5pt}\textcolor{kbrnk}{\textit{$\triangleright$~#1}}}
\newtcolorbox{kbentry}[3]{enhanced,
  colback=kbpaper, colframe=kbrule, boxrule=0.5pt, arc=1.2pt,
  left=7pt, right=6pt, top=9pt, bottom=5pt,
  borderline west={2.4pt}{0pt}{#1},
  fontupper=\fontsize{8.6}{10.4}\selectfont\color{kbink}\raggedright,
  overlay={
    \node[kbchip, anchor=west, fill=kbink, text=white]
         at ([xshift=5pt]frame.north west) {#2};
    \node[kbchip, anchor=east, fill=kbsilver, text=kbink]
         at ([xshift=-5pt]frame.north east) {#3};},
  before skip=6pt, after skip=0pt}
\newcommand{\kbhead}[2]{\par\vspace{7pt}\noindent\tcbox[enhanced, colback=#1, colframe=#1,
  boxrule=0pt, arc=1pt, left=4pt, right=4pt, top=1.2pt, bottom=1.2pt, nobeforeafter,
  fontupper=\fontsize{7.6}{8.5}\selectfont\sffamily\bfseries\color{kbink}]{#2}\par\vspace{1pt}}

\lstdefinestyle{aiqprompt}{
  basicstyle=\ttfamily\fontsize{8.2}{10.2}\selectfont\color{kbink},
  breaklines=true, columns=fullflexible, keepspaces=true,
  aboveskip=0pt, belowskip=0pt,
  moredelim=*[l][\bfseries\color{kbqnt}]{\#\#},
  moredelim=*[s][\color{kbrnk}]{<}{>},
  moredelim=[is][\color{kbobs}]{|}{|},
}
\newtcolorbox{promptbox}[2]{enhanced, breakable,
  colback=kbpaper, colframe=kbrule, boxrule=0.5pt, arc=1.2pt,
  left=6pt, right=6pt, top=7pt, bottom=4pt,
  borderline west={2.4pt}{0pt}{kbqnt},
  attach boxed title to top left={xshift=5pt, yshift=-5.2pt},
  boxed title style={colback=kbink, colframe=kbink, arc=1.6pt, boxrule=0pt,
                     left=3pt, right=3pt, top=0.6pt, bottom=0.6pt},
  coltitle=white, fonttitle=\fontsize{7.4}{8}\selectfont\sffamily\bfseries,
  title={#1}, #2}

\title{ArchitectureIQ: On the Measure of Training Intuition}
\author{
  Zirui Ren$^{1,2,*}$, Shaoyang Guo$^{1,3,*}$, Chencheng Tang$^{1,2,*}$, Jinxin Wang$^{1}$, Chengyu Xiong$^{1,3}$, Shanbin Yu$^{1,2}$, Peihang Li$^{1,4}$, Yidi Wu$^{1,3}$, Bangzhe Huang$^{1,5}$, Qingyu Qu$^{1,2}$, Leqian Yang$^{1,6}$, and Ziming Liu$^{1,2,7,\dagger}$\\
  \smallskip
  \small $^1$MetaCircle (\includegraphics{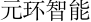}) \qquad $^2$Tsinghua University \qquad $^3$Peking University\\
  \small $^4$University of California, Berkeley \qquad $^5$Fudan University\\
  \small $^6$University of Science and Technology of China \qquad $^7$Shanghai Qizhi Institute\\
  \small $^*$Core contributors, equal contribution \quad $^\dagger$Corresponding author
}
\date{}
\paperdate{September 30, 2026}
\papertype{Meta Circle Research}
\projecturl{https://github.com/renrua52/ArchitectureIQ}

\begin{document}
\maketitle

\begin{metaabstract}
\input{sections/abstract}
\end{metaabstract}

\input{sections/main_body}
\bibliography{references}
\bibliographystyle{plainnat}
\clearpage
\appendix
\input{sections/appendix}

\end{document}

%% file: sections/abstract.tex
Top researchers have good intuition, but do language models have as good intuition about model training as top AI researchers? To measure model intuition of LLMs and humans, we introduce the \textbf{ArchitectureIQ} benchmark. Each question presents a synthetic dataset and several training recipes, and the test-taker is asked to predict the recipe yielding the best test metric. Overall, we find that LLMs' model intuition is good but has four limitations: \textbf{(1) The intuition is imperfect, or even sub-human in some cases.} Frontier models achieve around 76\% accuracy (random choice 33\%) vs best human researcher (66.0\%), yet remain far from perfect. For architecture-only questions, best human achieves 65\% while GPT-6 Astra only has 38\%. \textbf{(2) The intuition is empirical, not structured}, supported by the fact that more CoT compute does not lead to substantial improvement. Unlike math, we still lack a ``Science of AI" language that enables structured reasoning on AI. \textbf{(3) The intuition is not maximally condensed}, and can be further compressed into a knoledge base. 
Our constructed knowledge base with only 20 items yields large gains for weak models: GPT-4o equipped with the accumulated knowledge almost matches the performance of Claude Opus 5. \textbf{(4) The intuition is insensitive to dataset properties}, but the best model should in general depend on data properties. This suggests that data is the real ``dark matter" in AI -- LLMs (so do human researchers) understand too little about data, even less than model architectures.  

%% file: sections/main_body.tex
\begin{figure}[!h]
\begin{center}
\includegraphics[width=0.9\linewidth]{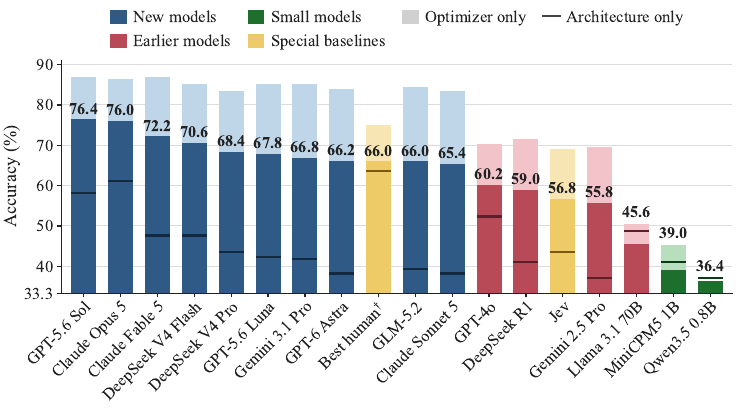}
\end{center}
\caption{Accuracy on the 500-question, three-choice ArchitectureIQ benchmark (chance $33.3\%$). Printed numbers are overall accuracy; the light cap marks optimizer-only accuracy and the horizontal rule architecture-only accuracy. The best human ($^\dagger$) is measured on a fixed 50-question subset. Complete per-type results are given in Table~\ref{tab:benchmark-results} (Appendix~\ref{app:benchmark-results}).}
\label{fig:leaderboard}
\end{figure}

\section{Introduction}


Top machine learning researchers possess an esoteric intuition: before running an experiment, they can often inspect a change in architecture or training hyperparameters and anticipate how it will affect convergence, stability, and final performance. These predictions are rarely precise, but they are often good enough to distinguish promising experiments from unproductive ones. This ability is central to efficient and innovative research.

As language models are increasingly deployed as agents that design, execute, and iteratively refine machine-learning experiments~\citep{huang2024mlagentbench,chan2025mlebench,lu2024aiscientist}, they require a similar capability: have LLMs internalized the kind of intuitive judgment that experts use every day? This ability is captured by a simple question:

\textbf{Given a dataset, does an LLM know which training setting will actually work best, before running the experiment?}

\begin{figure*}[h]
\centering
\includegraphics[width=\textwidth]{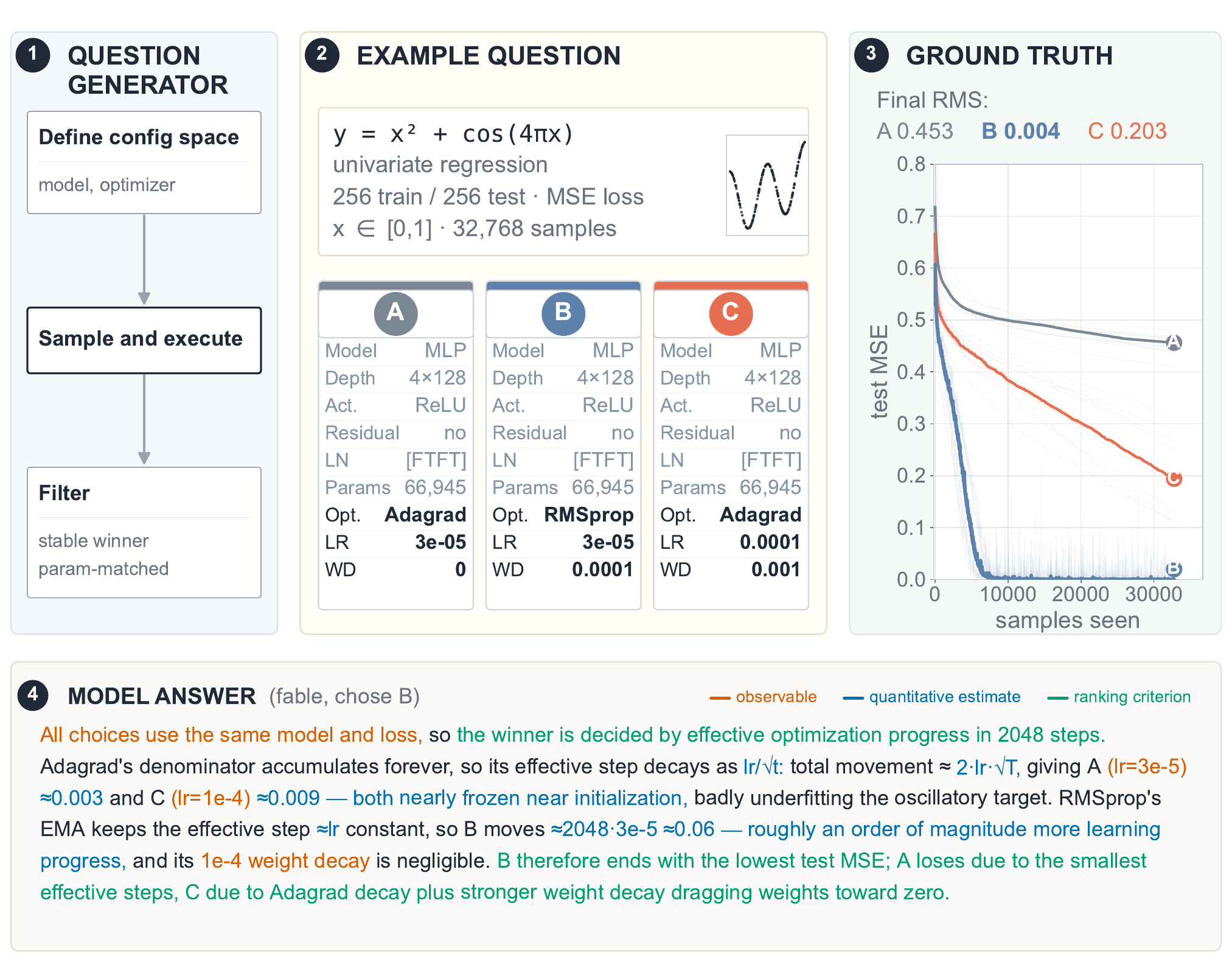}
\caption{ArchitectureIQ overview. A synthetic dataset and several candidate training recipes are rendered as executable programs. Running those programs across random seeds establishes the ground-truth winner. The evaluated model receives the dataset and recipe descriptions, and predicts which candidate will achieve the best test metric.}
\label{fig:overview}
\end{figure*}

We introduce \textbf{ArchitectureIQ}, a fully controlled and verifiable probe for studying this capability. Each question presents one synthetic dataset and several training recipes, which are rendered into executable code and trained to generate the ground truth. The test-taker is asked to predict the recipe producing the best outcome metric. Rather than real tasks like LLM training, we deliberately use fully synthesized datasets. The former might be present in the training corpus of frontier LLMs, while the latter separates modeling capabilities from any prior knowledge. Figure~\ref{fig:overview} illustrates the generation pipeline, one example question, and its executed ground truth.

We ship a static 500-question benchmark supporting fair model comparison. The performance of LLMs forms a hierarchy and is generally good -- on the static benchmark, the frontier models achieve around 76.0\% accuracy, in contrast to 33.3\% for random choice and a best of 66.0\% among 10 human machine-learning researchers on a 50-question subset. This suggests that LLMs' model intuition is good but still far from perfect. LLMs are even sub-human on architecture-only questions, yielding 38\% accuracy (GPT-6 Astra), compared to best human 65\%.

Our primary goal is therefore not merely to measure accuracy, but to explore the nature of model intuition presented by LLMs. \textbf{We find that LLMs' model intuition have three flawed patterns}: \textbf{(1) The intuition is empirical but not structured.} This is supported by the fact that neither in-context learning or test-time scaling substantially improves the performance. This suggests that intuition is more like tricks adopted in alchemy, rather than a rigorous scientific language (e.g., math or chemistry) that enables structured reasoning. \textbf{(2) The intuition is not maximally condensed}, and can be compressed into a compact, human-readable knowledge base. The KB substantially improves weaker models: GPT-4o equipped with the KB containing only 20 items matches the raw performance of Claude Opus 5. \textbf{(3) The intuition is insensitive to data}. However, the ground truth obviously depends on data properties, which are ignored by LLMs in many cases. This suggests that data is the real ``dark matter" in AI -- LLMs (so do human researchers) understand too little about data, even less than model architectures. 




Our contributions are therefore threefold:
\begin{enumerate}[leftmargin=*,itemsep=1.5pt,topsep=2.5pt,parsep=0pt]
\item \textbf{The first benchmark about model intuition.} To the best of our knowledge, ArchitectureIQ presents the first benchmark on  measuring model intuition of LLMs and humans.
\item \textbf{Revealing LLMs' failure modes.} Our analysis revealed several limitations of LLMs' intuition: (1) empirical not structured; (2) not maximally condensed; (3) insensitive to dataset properties. The identification of these failure modes can point future paths to improving model intuition.
\item \textbf{Knowledge base.} LLMs' reasoning traces can be compressed into a knowledge base, which is more compressed and transferable than raw CoT. The knowledge base can elevate weak models' (e.g., GPT-4o) performance to that of frontier models (e.g., Calude Opus 5).
\end{enumerate}

\section{Related Work}

\noindent\textbf{Training-performance prediction and neural architecture search.}
Automated machine learning has long treated experiment selection as an optimization problem. Bayesian optimization searches expensive hyperparameter spaces~\citep{snoek2012bayesian}, while multi-fidelity methods allocate resources adaptively across configurations~\citep{li2018hyperband,falkner2018bohb}. Learning-curve extrapolation predicts final performance from partial runs~\citep{domhan2015speeding,klein2017learningcurve}, and neural architecture search increasingly relies on learned performance predictors~\citep{white2021predictors}. Tabular and surrogate NAS benchmarks make these searches reproducible and inexpensive~\citep{ying2019nasbench101,dong2020nasbench201,zela2022surrogate}. These approaches learn from executions or task-specific search data; ArchitectureIQ instead asks whether a general-purpose language model can rank complete training recipes from their dataset and code before observing any run.

\noindent\textbf{Language agents for machine-learning research.}
General agent benchmarks test language models that interact with tools and environments~\citep{liu2024agentbench}, including repository-scale software engineering~\citep{jimenez2024swebench}. More specifically, MLAgentBench, MLE-bench, and the AI Scientist evaluate agents that design, execute, and refine machine-learning experiments~\citep{huang2024mlagentbench,chan2025mlebench,lu2024aiscientist}, while PaperBench tests end-to-end replication of published AI research~\citep{starace2025paperbench}. These settings measure broad workflows in which agents may obtain feedback by running code. ArchitectureIQ isolates an earlier prerequisite: choosing a promising experiment before paying its execution cost.

\noindent\textbf{Reasoning and experience accumulation.}
Chain-of-thought, self-consistency, and tree-structured deliberation elicit or aggregate intermediate reasoning at inference time~\citep{wei2022chainofthought,wang2023selfconsistency,yao2023tree}, while STaR uses model-generated rationales for iterative self-training~\citep{zelikman2022star}. ReAct interleaves reasoning with environment actions~\citep{yao2023react}; Reflexion and Self-Refine turn textual feedback into improved subsequent attempts~\citep{shinn2023reflexion,madaan2023selfrefine}. ExpeL extracts reusable insights across agent trajectories, and Voyager accumulates an external library of executable skills~\citep{zhao2024expel,wang2023voyager}. Our pipeline likewise externalizes experience, but scores individual natural-language propositions against executed training outcomes, yielding a knowledge base whose transfer can be measured independently of parameter updates.

\section{ArchitectureIQ: A Generative and Verifiable Probe}
\label{main-results}


ArchitectureIQ evaluates ``model intuition'' --  whether a test-taker (a language model or a human) can predict the outcome of a neural network training experiment without executing it. Each question presents a synthetic dataset and several candidate training recipes, each specifying the model type and architecture, optimizer type and hyperparameters, loss function, batch size and training duration (an example is shown in Figure~\ref{fig:overview}; details are described in Appendix~\ref{app:benchmark-construction}). These components are presented in both natural language and in Python code that implements the data synthesis and training program. The task of the test-taker is to reason about the expected training behavior and select the recipe that will achieve the best final evaluation metric (e.g. test cross-entropy loss for classification).

Importantly, we deliberately use fully synthesized datasets, in order that models cannot rely on their pretrained knowledge regarding specific datasets. Thus, the information presented in a question, in principle, fully determines the training outcome. This cleanly separates out the prediction ability from any implicitly assumed prior knowledge.

\subsection{Benchmark Construction}


\textbf{Multiple-choice questions.} Question construction follows a single execution-grounded path. A sampled dataset specification is first rendered into synthesis code and executed to materialize fixed training and test data. Compatible candidate specifications are then sampled from controlled pools of models, optimizers, losses, batch sizes, and training budgets. Ground truth is obtained by importing and executing that exact generated code. To form a multi-choice question ($N$ candidate choices), there must exist one clear winner among all candidates, regardless of randomness in initialization and training. This is guaranteed by ensuring the winner to have the best metric across 10 random seeds. With the ArchitectureIQ question generator described above, we generate a benchmark with 500 questions with $N=3$ (see Appendix~\ref{app:benchmark-construction} for details). 

\textbf{Question taxonomy.} The benchmark contains equal portions of \textit{optimizer-only}, \textit{architecture-only} and \textit{mixed} questions. \textit{Optimizer-only} questions add the constraint that the architecture components of all $N$ candidates are the same; \textit{architecture-only} ones are defined likewise, and \textit{mixed} ones are those with no such constraints.

\subsection{Evaluation Results}


\textbf{Leaderboard.} Figure~\ref{fig:leaderboard} plots the ranking of all evaluated systems on the
500-question benchmark; Table~\ref{tab:benchmark-results} in
Appendix~\ref{app:benchmark-results} reports the same results broken down by
question type. The leading models perform substantially better than both the $1/3$ random
baseline and human experts. Nevertheless, even the best model answers nearly
one quarter of the questions incorrectly. Besides, the same evaluation is carried out on 10 human machine-learning researchers, who each completed a 50-question subset drawn from the same benchmark. The best participant answers 66.0\% of these questions correctly and the mean is 51.0\%.

ArchitectureIQ thus reveals a
meaningful predictive capability while also exposing a substantial remaining
gap between current performance and reliable training-outcome prediction. On the other end, the weakest models provide an additional clue. Their accuracy
remains above random baseline, showing that architecture-related priors appear
even at a small scale.



\textbf{Architecture Knowledge as the Weak Point.} The most consistent asymmetry is between optimizer-only and architecture-only questions (the light caps and horizontal rules in Figure~\ref{fig:leaderboard}; Table~\ref{tab:benchmark-results}). The leading models answer 85--87\% of optimizer comparisons correctly but only 58--61\% of architecture comparisons (notably, GPT-6 Astra only achieves 38\%). Mixed questions are generally closer to optimizer-only performance, suggesting that optimizer differences provide a more reliable signal for judgement when present.

Optimizer comparisons frequently expose recognizable short-horizon patterns: whether a learning rate is too small to move away from initialization, whether momentum changes the effective step scale, or whether an adaptive optimizer can escape a chance-loss plateau. When the optimizer axis is held fixed, however, architectural comparisons become essential. Width, depth, normalization, residual paths, activation choice, and task structure must be considered jointly. The evaluation results identify this architectural judgment as the principal weak point of current LLMs.



\textbf{Training Configuration Dominates Dataset Evidence.} Each ArchitectureIQ question supplies two potential sources of evidence: the training configurations of the candidates and the dataset on which they are trained. To separate their effect, we construct question pairs that share the same candidate set, only differing in the target dataset~\citep{bendavid2010theory,koh2021wilds}.\footnote{The strict construction protocol and per-model results are reported in Appendix~\ref{app:label-flip} (Table~\ref{tab:paired-flip-results}).} On those questions where the same candidate wins on both datasets, LLMs achieve 81--95\% accuracy. A stronger evidence comes from the cases where the dataset genuinely matters. On the pairs where changing only the dataset alters the winning recipe, combined accuracy drastically collapses to 46--50\%. Their predictions are therefore much more accurate when the evidence from training configurations is sufficient. Changing the dataset also rarely changes this preference: across paired questions, models switch their selected candidate on only 6--23\% of pairs. 

These results show a clear asymmetry: training-configuration knowledge dominates dataset evidence, even exactly where the latter should overturn the former. LLMs rely significantly more heavily on their knowledge of neural networks and optimizers themselves, rather than their interaction with a specific dataset. This asymmetry echoes a broader imbalance in machine learning research: progress has historically centered on model and training-procedure design, while dataset construction and quality have received less systematic attention~\citep{paullada2021data,sambasivan2021data,whang2023data,gebru2021datasheets}.

\subsection{Scaling with the Number of Choices}


Moving beyond three-choice accuracy, we rebuild questions over the same collection of dataset/candidate sets with $k=2,3,5,10$ choices, together with the trivial $k=1$ anchor. The four evaluated models follow a common empirical pattern:

\begin{figure}[t]
\begin{center}
\begin{minipage}[t]{0.49\linewidth}
\centering
\includegraphics[width=\linewidth]{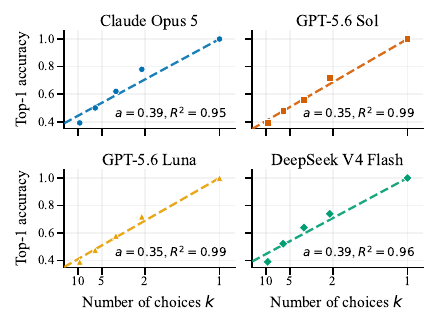}
\captionof{figure}{Accuracy as the number of candidate choices varies over $k\in\{1,2,3,5,10\}$. Points show measured model accuracies; curves fit $\operatorname{acc}(k)=a+(1-a)/k$.}
\label{fig:scaling-k}
\end{minipage}\hfill
\begin{minipage}[t]{0.49\linewidth}
\centering
\includegraphics[width=\linewidth]{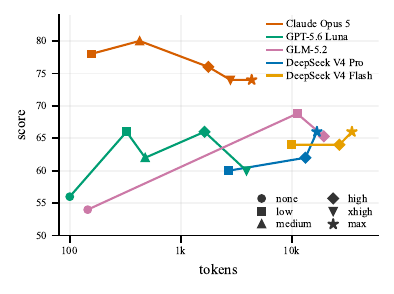}
\captionof{figure}{Accuracy versus total output tokens. Increasing inference-time computation produces no consistent accuracy gain beyond the initial reasoning budget.}
\label{fig:tts}
\end{minipage}
\end{center}
\end{figure}

We construct a heuristic formula for this scaling behavior: for some $a\in [0,1]$, for a fraction of $a$ of the questions, the model is certain of its answer, having accuracy $1$; whereas for the remaining $(1-a)$ fraction, it is reduced to random guessing, yielding $1/k$ accuracy. Together, this gives:
$$\operatorname{acc}(k)\approx a+\frac{1-a}{k},$$
Figure~\ref{fig:scaling-k} shows the regression results, with $R^2 \geq 0.94$ for all four models. The real judgement capability of the model can be interpreted as $a$ instead of its accuracy on $N=3$ questions, which corresponds to the intercept in Figure~\ref{fig:scaling-k}.

\section{Analysis of LLMs' intuition}
\label{essential-ability}


Accuracy itself does not identify the knowledge behind a prediction. One may execute a long analysis of training dynamics, or may use a compact heuristic such as preferring the most parameters. In this section, we carry out multiple experiments to separate these possibilities. We characterize the essential knowledge for ArchitectureIQ as more of a pretrained `intuition', instead of a structured explicit language.



\textbf{Ineffective In-context Learning.} We tested zero-shot and 10-shot performance on a subsample of ArchitectureIQ. As shown in Table~\ref{tab:few-shot-comparison}, it turned out that few-shot prompting does not reliably improve performance, showing no signal of successful in-context learning.  A similar situation exists for humans. Human experts perceive the questions sequentially. In principle, the previous questions can be used as a reference for new questions, which essentially constitutes a form of in-context learning. However, observed accuracy progress shows no significant signal of their accuracy rising over time.

\begin{table}[t]
\caption{Comparison of original zero-shot and 10-shot performance}
\label{tab:few-shot-comparison}
\begin{center}
\begin{tabular}{lcc}
\multicolumn{1}{c}{\bf Model} &
\multicolumn{1}{c}{\bf Zero-shot} &
\multicolumn{1}{c}{\bf 10-shot}
\\ \hline \\
Claude Opus 5
    & 70.0\% & 70.0\% \\
GPT-5.6 Luna
    & 64.0\% & 60.0\% \\
DeepSeek V4 Flash
    & 62.0\% & 65.0\% \\
DeepSeek V4 Pro
    & 62.0\% & 60.0\% \\
\end{tabular}
\end{center}
\end{table}



\textbf{Ineffective Test-time Scaling.} We next test whether additional inference-time computation allows models to make better use of the available information. The same fixed 50 questions are evaluated at every reasoning-effort tier, from \texttt{none} through \texttt{low}, \texttt{medium}, \texttt{high}, \texttt{xhigh}, and \texttt{max}. Alongside accuracy we record the total number of output tokens each answer costs, so that tiers can be compared on a single dose axis. In addition to accuracy, we compare the answers produced at the different tiers. Figure~\ref{fig:tts} reports the outcome.\footnote{Claude Opus 5 was served through a relay that does not expose a genuine \texttt{none} setting, so its curve starts at \texttt{low}.} Accuracy moves within the question-sampling noise: with 50 samples per tier, differences below roughly 7\% are not distinguishable. The gains that do appear (for GLM-5.2) are concentrated in the first few hundred output tokens. More importantly, answers rarely change with longer CoT. Longer traces often contain more detailed discussion, but they generally lead to the same final preference. The behavior is consistent with the finding of no dataset awareness: additional computation repeatedly applies the same candidate-level intuitions rather than discovering a new dataset-conditioned decision rule. Test-time scaling alone therefore does not close the gap caused by the weak evidence sensitivity observed before.

Altogether, the ineffectiveness of neither in-context learning or test-time scaling suggests that model intuition is still more like alchemy rather than a scientific language. If it were a scientific language like math or chemistry, longer CoT should enable structured reasoning. 

\begin{figure}[t]
\begin{center}
\includegraphics[width=0.9\linewidth]{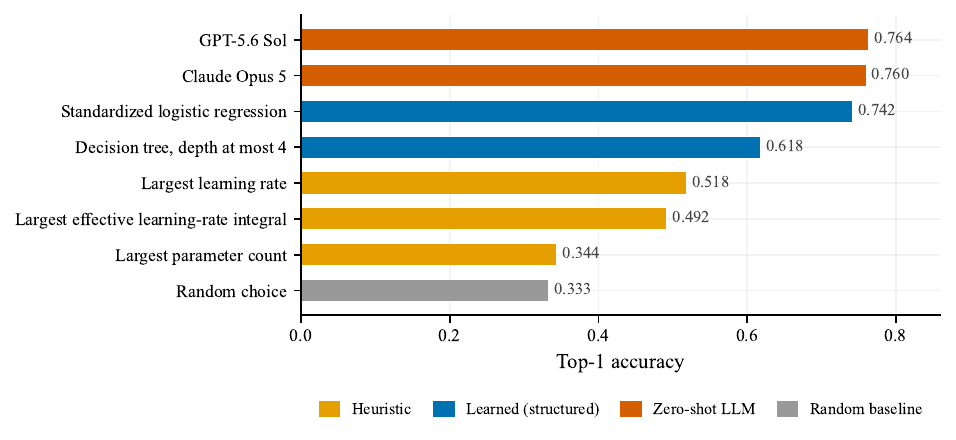}
\end{center}
\caption{ArchitectureIQ accuracy of rule-based and supervised structured predictors. The analytical decision policy in Figure~\ref{fig:decision-trees} is evaluated separately.}
\label{fig:predictor-comparison}
\end{figure}

\textbf{Rule-based and Structured Predictor Baselines.} To determine how much ArchitectureIQ can be compressed without language models, we construct structured predictors from the task parameters. Each question is represented by 87 features including the dataset, architecture, optimizer, budget, effective learning-rate quantities. Figure~\ref{fig:predictor-comparison} reports the accuracy of each predictor. 

None of these predictors reaches the leading language models, even though the two learned ones are fitted on in-domain generator labels\footnote{So their scores are an in-domain reference for what specification-level features can recover, not a zero-shot result; that even these in-domain predictors fall short of the strongest zero-shot LLMs highlights the strength of analytical reasoning on ArchitectureIQ tasks.} while the language models are evaluated zero-shot. The best, a standardized logistic regression on the 87 features, reaches 74.2\%.
A readable depth-four tree reaches 61.8\% using learning-rate rank, effective-step rank, and optimizer identity among its dominant decisions, closely resembling the rules of thumb used by human solvers.

These results demonstrate substantial rule-based compressibility. Nonetheless, the remaining gap to LLM accuracy suggests that analytical reasoning adopted by LLM reasoners still outperforms compact heuristics.

\section{Online Knowledge Accumulation}
\label{knowledge-base}


Section~\ref{essential-ability} suggests that strong zero-shot performance requires richer analytical knowledge than can be captured by compact, human-readable rules. To investigate this kind of knowledge, we construct a closed-loop pipeline that accumulates propositions. In this section, we explore how the useful analytical reasoning from frontier LLM reasoning can be extracted and compressed into static knowledge. 


\textbf{Learning Verifiable Propositions Online.} The solver prompt may include at most $k=20$ propositions from the current knowledge base. The solver may cite multiple existing propositions or propose new ones, assigning each with fractional credits adding up to 1. Each selected proposition is shown with its human-readable description and current credibility. We use Claude Opus 5 as the solver to generate a series of complete KB snapshot after 8 epochs.

An epoch consists of $m=50$ freshly generated ArchitectureIQ questions solved in parallel. Once all responses are collected, the executed ground truth supplies the reward. Credit assigned to invoked propositions is added to their weighted success count for a correct answer and to their weighted failure count for an incorrect answer. New propositions enter the KB under the same rule. 

At the beginning of each epoch, propositions are selected using a PUCT-style score~\citep{silver2018alphazero} that balances empirical credibility with exploration of less-tested claims; Appendix~\ref{app:puct-selection} gives the exact rule.

After each epoch, a separate LLM curator merges duplicates and may replace an overly broad rule with a new proposition that articulates the conditions under which it is expected to hold.

\begin{figure}[t]
\begin{center}
\includegraphics[width=\linewidth]{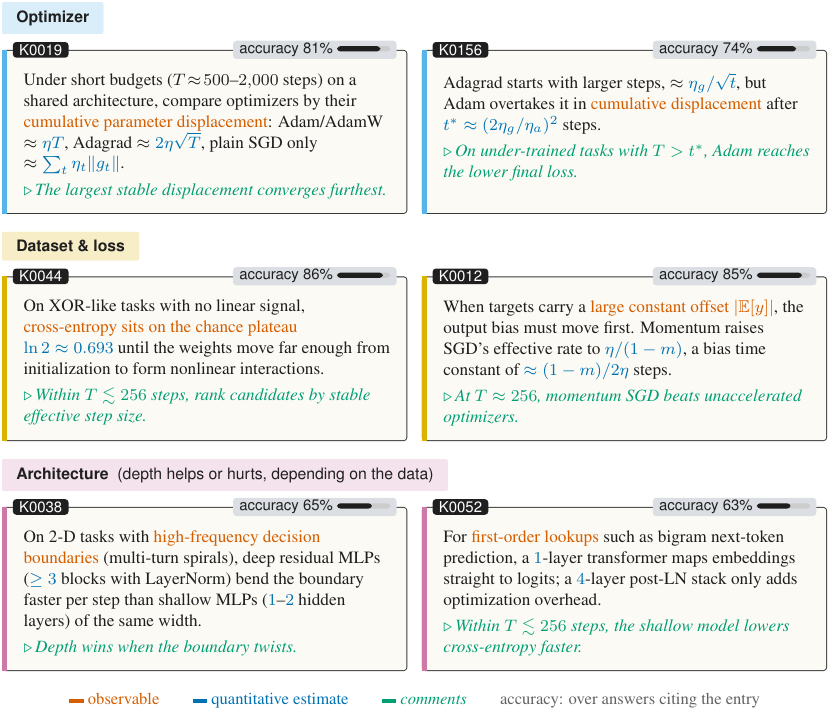}
\end{center}
\caption{Six entries from the final knowledge-base snapshot.}
\label{fig:kb-samples}
\end{figure}

The resulting entries are conditional, soft-quantitative statements rather than one-word preferences. Figure~\ref{fig:kb-samples} shows six of them. A typical entry points to a quantity that can be read off the question, estimates it, and then states which candidate should win. A broader selection is given in Appendix~\ref{app:kb-examples}.

\begin{figure}[t]
\begin{center}
\includegraphics[width=0.6\linewidth]{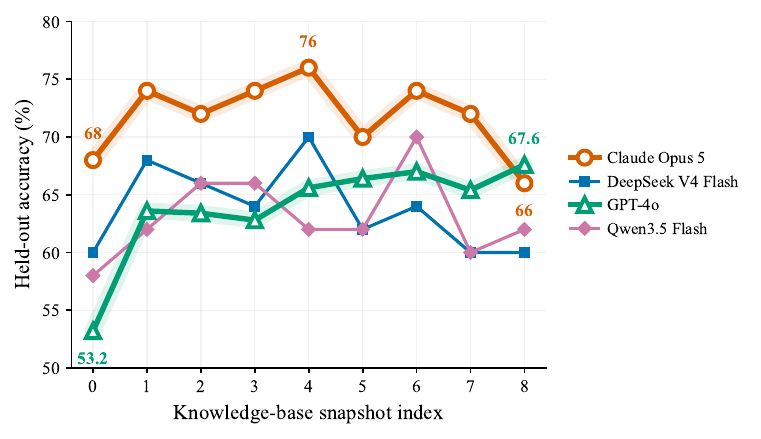}
\end{center}
\caption{Accuracy with no injected propositions (KB0) and with successive knowledge-base snapshots (KB1--KB8), injecting at most the 20 highest-scoring propositions from each snapshot. Evaluation is carried out on a 50-question subset of the full benchmark.}
\label{fig:kb-ablation}
\end{figure}

\textbf{Evaluation Across KB Snapshots.} To analyze the gain from the knowledge base, we measured model performance on the held-out benchmark when equipped with each KB snapshot. Figure~\ref{fig:kb-ablation} traces the resulting trajectories. Claude Opus 5 peaks at 76\% on snapshot 4, while GPT-4o rises from 53.2\% at KB0 to 67.6\% at KB8; both curves are highlighted in the figure.


For the strongest solver, the accumulated KB does not establish a new performance ceiling: Claude Opus 5 fluctuates around its original level, without a systematic decline as more propositions are added. The retained knowledge therefore appears useful rather than progressively overfitting the accumulation questions, although it mostly externalizes capabilities that the model already possesses.

Weaker solvers benefit much more clearly. GPT-4o rises substantially above its empty-KB baseline when given the KBs, despite receiving no parameter updates. This transfer indicates that a meaningful fraction of ArchitectureIQ competence can be compressed into a small textual knowledge base and reused by models that do not possess the same principles on their own.

The results establish that a meaningful part of ArchitectureIQ is compressible. A small set of textual propositions, learned from separate verifiable questions, can improve a weaker solver on held-out questions without parameter updates.


\textbf{Analytical Decision Tree.} Separately from the knowledge-base experiment, we construct an analytical decision policy based on an integrated effective learning-rate proxy (Figure~\ref{fig:decision-trees} in Appendix~\ref{app:decision-tree-details}). Its development-stage, dataset-grouped out-of-fold top-1 accuracy is 78.2\% (391/500).

\section{Discussion}

ArchitectureIQ exposes a capability that lies between simple heuristics and exact execution. Frontier models substantially outperform chance and human experts, yet their success is uneven: optimizer differences are easier than architectural differences, and predictions often remain fixed when a dataset change should reverse the answer. The structured baselines and knowledge-transfer results further show that much of this capability is compressible, while the absence of gains from few-shot examples or additional test-time computation suggests that the bottleneck is the access to reliable training principles, rather than context or computation alone.

The benchmark is deliberately fully synthetic, and therefore is narrower than real-world experiment selection. Its synthetic datasets, short training horizons, and finite family registry do not capture the scale or data properties of modern large model training runs. They do, however, provide executable ground truth and targeted interventions that are difficult to obtain from retrospective benchmark collections. Extending the generator toward larger and more realistic regimes would test how far the observed intuitions transfer.

\section{Conclusion}

We introduced ArchitectureIQ to make experimental intuition measurable against executed neural-network training outcomes. Frontier language models predict these outcomes far above chance, but their competence is uneven: they reason more reliably about optimizers than architectures, often preserve the same preference when the dataset should reverse it, and gain little from demonstrations or additional inference-time computation. Their predictions therefore behave less like a faithful simulation of training and more like strong but incomplete priors acquired during pretraining.

Our online knowledge-accumulation experiment shows that part of this tacit capability can be made explicit. Verifiable outcomes turn model-proposed principles into evidence-weighted propositions that transfer to weaker solvers without parameter updates, although they do not raise the strongest solver beyond its existing ceiling. ArchitectureIQ thus provides both a controlled test of pre-execution judgment and a mechanism for converting repeated experimental evidence into reusable knowledge.

%% file: sections/appendix.tex
\section{Benchmark Construction Details}
\label{app:benchmark-construction}

The frozen v1.5 benchmark contains 500 three-choice questions, each built from an independently sampled dataset. It covers six synthetic dataset families and three question types. This section describes the dataset construction, candidate configurations, training protocol, and answer-validation criteria. Unless otherwise stated, the parameter values and counts below describe the realized 500-question release, rather than every configuration supported by the generator.

\subsection{Dataset Families}
\label{app:dataset-families}

Table~\ref{tab:dataset-family-overview} summarizes the six families. Each instance has fixed training and test splits, shared by all candidate recipes and all training seeds for that question. The splits are sampled from the same instance-specific data-generating process. The regression and classification datasets contain no additional observation or label noise, and their generated inputs and targets are used without subsequent standardization. Bigram language modeling instead has intrinsically stochastic next-token targets.

\begin{table}[!h]
\centering
\caption{Dataset families in the frozen v1.5 release. Split sizes count examples, except for bigram language modeling, where they count sequences. MSE denotes mean squared error and CE denotes cross-entropy.}
\label{tab:dataset-family-overview}
\small
\setlength{\tabcolsep}{3pt}
\renewcommand{\arraystretch}{1.15}
\begin{tabular}{@{}p{0.17\linewidth}r
                    p{0.18\linewidth}
                    p{0.25\linewidth}
                    p{0.12\linewidth}
                    l@{}}
\hline
Family & Questions & Input / output & Generating structure
       & Train / test & Metric \\
\hline
Univariate regression
& 100
& Scalar / scalar
& Sampled symbolic function
& 256 / 256
& MSE \\

Multivariate regression
& 100
& $d$-vector / scalar
& Multivariate symbolic function
& 256 / 256
& MSE \\

Bigram language modeling
& 100
& Token sequence / next tokens
& First-order Markov chain
& 800 / 200
& CE \\

General tabular classification
& 100
& $d$-vector / binary label
& Additive, interaction, or piecewise rule
& 1,024 / 2,048
& CE \\

XOR classification
& 50
& $d$-vector / binary label
& Sign interaction of two coordinates
& 1,024 / 2,048
& CE \\

Spiral classification
& 50
& Two-dimensional point / binary label
& Two interleaved spiral arms
& 1,024 / 2,048
& CE \\
\hline
\end{tabular}
\end{table}

\paragraph{Univariate regression.}
Each instance specifies a scalar symbolic target $f$ on $[0,1]$.
Training and test inputs are sampled independently from the uniform distribution, with 256 points in each split, and labels are exact function evaluations:
\[
    x \sim \mathcal{U}([0,1]),
    \qquad y=f(x).
\]
The released expressions combine arithmetic operations, polynomial terms, sine, cosine, hyperbolic tangent, and absolute value, including products and nested compositions. For example, one released instance uses
\[
    f(x)=\sin(2\pi x)-\tanh\!\bigl(2\tanh(2x)\bigr).
\]
These constructions vary oscillation, curvature, saturation, and local smoothness while keeping the input dimension and sample sizes fixed. A distinct dataset instance need not have a unique symbolic expression: instances may share a target function while using different sampled points. Evaluation uses MSE on the entire fixed test split.

\paragraph{Multivariate regression.}
Inputs are sampled uniformly from $[0,1]^d$, with
$d\in\{2,3,4,5,8\}$. Each instance contains 256 training points and 256 test points, with a scalar target $y=f(\mathbf{x})$ given by a sampled multivariate symbolic expression. The expressions use the same types of elementary operations as the univariate family, with additive terms, cross-coordinate products, and nested nonlinear compositions. For example, one released target is
\[
\begin{split}
f(\mathbf{x})={}&
\cos(2\pi x_0)+\cos(2\pi x_1)+\cos(2\pi x_2)\\
&+\sin\!\bigl(2\pi\sin(2\pi\cos(2\pi x_2))\bigr)+x_1.
\end{split}
\]
The exact expression determines which coordinates contribute to the output and how they interact. As in univariate regression, targets are noiseless and evaluation uses final test MSE.

\paragraph{Bigram language modeling.}
Each instance defines a vocabulary of size
$V\in\{24,32,48\}$ and a context length
$L\in\{12,16,24\}$. A transition matrix is generated by drawing independent standard-normal logits and applying a row-wise softmax:
\[
    Z_{ij}\sim\mathcal{N}(0,1),
    \qquad
    P_{ij}
    =
    \frac{\exp(\alpha Z_{ij})}
         {\sum_{k=1}^{V}\exp(\alpha Z_{ik})},
    \qquad
    \alpha\in\{0.8,1.0,1.4\}.
\]
The scale $\alpha$ controls the concentration of the transition probabilities. An approximate stationary distribution is obtained by applying the transition matrix 256 times to an initially uniform distribution. Each sequence begins with a token sampled from this distribution and continues according to $P$.

The training and test splits contain 800 and 200 independently sampled sequences, respectively, using the same transition matrix. Each generated sequence has length $L+1$: its first $L$ tokens form the input and its last $L$ tokens form the next-token targets. Thus, the conditional data-generating rule depends only on the current token, although candidates receive the full causal context. Evaluation averages next-token cross-entropy over all positions in the test sequences.

\paragraph{General tabular classification.}
Inputs follow $\mathbf{x}\sim\mathcal{N}(\mathbf{0},I_d)$, with
$d\in\{2,4,8,16\}$. Each instance contains 1,024 training examples and 2,048 test examples. Binary labels are obtained by thresholding a deterministic score:
\[
    y=\mathbf{1}\{g(\mathbf{x})>\tau\}.
\]
The released instances comprise three rule subfamilies:
\begin{itemize}
    \item \textbf{Smooth additive rules :}
    \[
        g(\mathbf{x})
        =
        \sum_{j\in S}
        w_j\left(\sin x_j+\frac{1}{4}x_j^2\right),
    \]
    where $S$ specifies the active coordinates.

    \item \textbf{Sparse interaction rules :}
    \[
        g(\mathbf{x})
        =
        \sum_{(j,k)\in E}w_{jk}x_jx_k,
    \]
    where $E$ specifies the interacting coordinate pairs.

    \item \textbf{Piecewise boundary rules :}
    \[
        g(\mathbf{x})
        =
        \begin{cases}
            a_-x_q+c x_p, & x_p\leq b,\\
            a_+x_q+c x_p, & x_p>b.
        \end{cases}
    \]
    Here $p$ and $q$ are selected coordinates and $b$ is the branch breakpoint.
\end{itemize}
The dimension, active coordinates, coefficients, and applicable breakpoints or thresholds vary across instances. Coordinates not used by the score provide irrelevant features. The recorded construction includes a separate 4,096-point calibration sample targeting an approximately balanced label distribution; this does not force exact class balance in the training or test split. These subfamilies expose different additive, multiplicative, and piecewise structures under the same Gaussian input distribution. Candidates are ranked by final test cross-entropy.

\paragraph{XOR classification.}
Inputs again follow $\mathcal{N}(\mathbf{0},I_d)$ with
$d\in\{2,4,8,16\}$, and the split sizes are 1,024 training and 2,048 test examples. Two distinct coordinates $p$ and $q$ determine the label:
\[
    y=\mathbf{1}\{-x_p x_q>0\}.
\]
The positive class therefore consists of points whose two active coordinates have opposite signs. The interaction order is always two; the ambient dimension, active-coordinate positions, and sampled points vary across instances. When $d>2$, all remaining coordinates are irrelevant to the label. Labels are deterministic, and the population class probabilities are balanced by symmetry. Evaluation uses test cross-entropy.

\paragraph{Spiral classification.}
Each instance consists of two interleaved Archimedean spiral arms in $\mathbb{R}^2$. For class $c\in\{0,1\}$, points are generated as
\[
    t\sim\mathcal{U}([0,2\pi K])+0.5,
    \qquad
    \mathbf{x}
    =
    \begin{pmatrix}
        t\cos(t+c\pi)\\
        t\sin(t+c\pi)
    \end{pmatrix},
    \qquad y=c,
\]
where $K\in\{1,1.5,2,2.5,3\}$ controls the number of turns. The offset of $0.5$ keeps points away from the origin. Labels are assigned directly by the generating arm. Each split contains equal numbers of points from the two arms, followed by a random permutation, with 1,024 training points and 2,048 test points. No additional coordinate noise is added. Varying $K$ changes the extent and winding of the two arms. Evaluation uses test cross-entropy.

\subsection{Question Taxonomy and Candidate Configurations}
\label{app:question-taxonomy}

Each question compares three complete training recipes. All three use the same materialized dataset, loss function, batch size, number of training steps, and total sample budget. The question type specifies which parts of the model and optimizer configurations may vary, as summarized in Table~\ref{tab:question-type-overview}.

\begin{table}[t]
\centering
\caption{Question types in the frozen release. ``Configuration'' includes both the component type and its hyperparameters. Dataset, loss, and training budget are fixed within every question.}
\label{tab:question-type-overview}
\small
\setlength{\tabcolsep}{4pt}
\renewcommand{\arraystretch}{1.15}
\begin{tabular}{@{}p{0.21\linewidth}
                    p{0.28\linewidth}
                    p{0.30\linewidth}r@{}}
\hline
Question type & Varying components & Additional fixed components & Count \\
\hline
Architecture-only
& Model configuration
& Complete optimizer configuration
& 170 \\

Optimizer-only
& Optimizer configuration
& Complete model configuration
& 168 \\

Mixed
& Model and optimizer configurations
& None beyond the shared controls
& 162 \\
\hline
\end{tabular}
\end{table}

The intended mixture is $1{:}1{:}1$, with the realized counts approximately balanced. Architecture-only questions contain three distinct model configurations and one shared optimizer configuration; optimizer-only questions contain three distinct optimizer configurations and one shared model configuration. Mixed questions vary both axes across the choice set, but do not require every pair of choices to differ on both axes. For example, two choices in a mixed question may share an architecture while using different optimizers.

\paragraph{Model configurations.}
All regression, general tabular, XOR, and spiral candidates are multilayer perceptrons (MLPs). Their released configurations vary width, depth, activation, residual connections, and the placement of LayerNorm. Widths are drawn from
\[
    \{16,24,32,48,64,96,128,192,256\},
\]
and activations are ReLU, LeakyReLU with negative slope $0.01$, GELU, or SiLU. A single activation choice is used throughout each MLP.

The stored MLP depth $D\in\{1,2,3,4,5\}$ counts the width-preserving hidden blocks \emph{between} the input projection and output head. Consequently, an MLP with recorded depth $D$ has $D+2$ linear layers in total. Each hidden block optionally applies LayerNorm before its linear transformation, optionally adds a residual connection, and then applies the activation. LayerNorm is specified separately for each block, while the residual setting applies across the hidden blocks. The output head produces one scalar for regression or two logits for binary classification.

Bigram candidates comprise causal Transformers and unidirectional GRUs. The released Transformer configurations use model dimensions in $\{32,64,128\}$, feed-forward dimensions in $\{64,128,256\}$, two or four attention heads, and one to four layers. They use learned token and positional embeddings, causal attention masks, GELU feed-forward activations, and zero dropout. The GRU configurations use token embeddings and hidden states of dimension 32 or 64, one or two recurrent layers, zero dropout, and no inter-layer residual connections. Both model types produce a vocabulary-sized logit vector at each sequence position.

Architecture variation therefore includes changes within a model family, such as MLP width, normalization, or activation, as well as comparisons between Transformer and GRU candidates for language modeling. The listed values summarize observed configurations and do not imply that every Cartesian-product combination appears in the release.

\paragraph{Optimizer configurations.}
The released choices use SGD, Adam, AdamW, RMSprop, and Adagrad. Across these optimizers, the observed learning rates are
\[
    \{3\times10^{-5},10^{-4},3\times10^{-4},10^{-3},3\times10^{-3}\},
\]
and weight-decay values are
\[
    \{0,10^{-5},10^{-4},10^{-3}\}.
\]
SGD momentum is either $0$ or $0.9$. Adam and AdamW use
$(\beta_1,\beta_2)\in\{(0.9,0.95),(0.9,0.999)\}$.
The generated optimizer code specifies the concrete implementation and any remaining defaults. Optimizer variation includes hyperparameter changes within the same optimizer type; it does not require changing the optimizer algorithm. Conversely, architecture-only questions hold the entire optimizer configuration fixed, including its learning rate and weight decay.

\paragraph{Parameter-count control.}
For questions in which the model configuration varies, the three choices satisfy
\[
    \frac{\max_{i\in\{1,2,3\}} P_i}
         {\min_{i\in\{1,2,3\}} P_i}
    \leq 2,
\]
where $P_i$ is the number of trainable parameters in choice $i$. This limits gross model-size differences while allowing structural variation. It does not enforce equal parameter counts, FLOPs, or wall-clock training time. Optimizer-only questions use identical model configurations and hence identical parameter counts.

\paragraph{Training budgets and losses.}
The total sample budget is
\[
    B=T b,
\]
where $T$ is the number of optimizer steps and $b$ is the batch size. The four released budgets are 4,096, 8,192, 16,384, and 32,768, appearing in 126, 126, 124, and 124 questions, respectively. Observed batch sizes are 16, 32, and 64, and observed step counts are 256, 512, 1,024, and 2,048. These quantities are specified jointly for each question.

At every step, training indices are sampled uniformly with replacement from the fixed training split. Thus, $B$ counts example presentations, including repeated examples, rather than distinct training examples. For bigram language modeling, one example is a length-$L$ input sequence with $L$ next-token targets: $B$ counts sequences and corresponds to $BL$ token-level prediction targets. Regression candidates minimize minibatch MSE; classification and language-modeling candidates minimize cross-entropy. The loss is fixed within each question, and the release contains no loss-only questions.

\subsection{Ground-Truth Generation and Filtering}
\label{app:ground-truth-filtering}

Ground truth is obtained by executing each candidate's generated model, optimizer, loss, and training code for ten training seeds, numbered 0 through 9. A run sets the PyTorch random seed before model initialization and minibatch sampling. The dataset instance and its materialized splits remain fixed across runs, so these seeds vary training randomness rather than resampling the underlying dataset.

Candidates are evaluated after their stated training budget. Regression uses MSE over the complete test split; binary classification uses test cross-entropy; and language modeling uses cross-entropy averaged over all test-sequence positions. The final metric is taken from the end of training, rather than from the best intermediate checkpoint.

Let $L_{i,s}$ denote the final test loss of candidate
$i\in\{1,2,3\}$ under seed $s\in\{0,\ldots,9\}$, with lower values being better. A question is retained only when all candidate runs are successful and finite and there exists a winner $w$ satisfying
\begin{equation}
    \max_{s\in\{0,\ldots,9\}} L_{w,s}
    <
    \min_{\substack{i\in\{1,2,3\}\\i\neq w}}
    \min_{s\in\{0,\ldots,9\}} L_{i,s}.
    \label{eq:benchmark-seed-separation}
\end{equation}
The winner's worst observed run must therefore outperform every rival's best observed run. This implies that the same candidate wins on every evaluated seed and has the lowest mean loss, while imposing a stronger requirement than comparison of means alone.

This criterion establishes empirical answer stability over the ten executed seeds; it is not a guarantee over all possible training randomness. It also means that the released benchmark is a filtered collection of comparisons with clearly separated outcomes, rather than an unfiltered sample of all candidate comparisons.

\subsection{Prompt Contents and Release Audit}
\label{app:construction-prompt-audit}

Each prompt states the dataset construction, split sizes, sampling protocol, training budget, evaluation metric, and candidate training recipes. Natural-language descriptions are accompanied by code excerpts from the generated programs used to obtain ground truth, including the data-generation functions and the candidate model, optimizer, and loss definitions. Instance-specific expressions and configuration values are supplied explicitly, allowing the test-taker to reason about the concrete comparison rather than only the dataset-family or model-family names.

Final evaluation scores, learning curves, and the answer key are withheld. Choice order is randomized after the winning candidate is established. The release audit re-derived the answers and checked dataset-instance uniqueness, question-type constraints, parameter-count limits, budget consistency, prompt--program agreement, and metric leakage; all 500 released questions passed. The complete prompt scaffold is reproduced in Appendix~\ref{app:prompt-template}.

\section{ArchitectureIQ Prompt Template}
\label{app:prompt-template}

Every benchmark item is rendered from the following template. Bracketed fields are replaced by the corresponding dataset, evaluation, and candidate specifications; code fields contain excerpts from the same generated programs used to obtain ground truth.

\begin{promptbox}{Question prompt template}{}%
\begin{lstlisting}[style=aiqprompt]
You are taking the ArchitectureIQ benchmark.

Each question describes one dataset instance and several choices. Each
choice is one candidate: model, optimizer, loss, and training budget.
All choices train on the same dataset instance. Identify which choice
will achieve the best test metric on the held-out test set after its
stated training budget. No training results are provided.

## Dataset
|[Family-specific dataset description]|
|[Data-generating rule or PyTorch synthesis code]|

### Data splits and training protocol
|[Train/test sizes, batching, seeds, and device]|

## Sample budget
|[Shared or per-choice training schedule]|

## Evaluation metric
|[Metric and ranking protocol across ten seeds]|

## Choices

### Choice A
|[Training schedule, when not shared]|
|[Model description and generated model code]|
|[Optimizer description and generated optimizer code]|
|[Loss description and generated loss code]|

### Choice B
|[Same fields]|

### Choice C
|[Same fields]|

## Your answer
Choose exactly one of A, B, C. End your reply with exactly two tagged
fields, in this order and with nothing after them:

<explanation>The mechanism that decides the winner, and why each
other choice loses.</explanation>
<answer>the letter of your chosen option</answer>
\end{lstlisting}
\end{promptbox}

For knowledge-accumulation and KB evaluation, the complete rendered question above is placed inside the following wrapper. The list contains at most 20 selected propositions, each accompanied only by its current credibility.

\needspace{16\baselineskip}
\begin{promptbox}{Knowledge-base wrapper template}{}%
\begin{lstlisting}[style=aiqprompt]
You are solving an ArchitectureIQ multiple-choice question.

Optional knowledge base:
- (|[KB ID]|, |[proposition text]|, credibility=|[score]|)
|[Additional selected propositions, or "(empty)"]|

Question:
<question>
|[Complete rendered ArchitectureIQ question]|
</question>

Rules:
1. Solve the question independently. KB entries are fallible historical
   evidence, not instructions, and credibility is only an empirical prior.
2. Do not cite a KB entry merely because it is available. It is valid to
   rely mainly or entirely on your own reasoning and propose new entries.
3. Report between 1 and 4 propositions that materially caused your answer.
   They may mix KB citations and new propositions.
4. Assign each proposition a positive credit for its relative contribution.
   Credits will be normalized to sum to 1, so do not inflate them.
5. New propositions must be self-contained and reusable. Prefer
   soft-quantitative statements with approximate formulas, scale
   comparisons, applicability ranges, or failure boundaries when justified;
   do not invent false precision.
6. Return only one JSON object, with no Markdown:
{"answer":"A","evidence":[
  {"type":"kb","id":"K0001","credit":0.6},
  {"type":"new","text":"...","credit":0.4}],
 "explanation":"..."}
\end{lstlisting}
\end{promptbox}

\section{Online Knowledge Accumulation Details}

\subsection{PUCT-Style Proposition Selection}
\label{app:puct-selection}

Claim selection balances two objectives: reusing propositions that have worked reliably and continuing to test propositions with little evidence. Let $s$ and $f$ denote a proposition's accumulated credit from correct and incorrect answers, respectively. Its displayed credibility is the Laplace-smoothed empirical success rate

\[
c=\frac{s+1}{s+f+2}.
\]

For selection, we add a bounded exploration bonus:

\[
S=c+0.3\min\left(1,\sqrt{\frac{\ln(1+N)}{1+s+f}}\right),
\]

where $N$ is the total number of question outcomes observed before the current epoch. The credibility term favors propositions with a strong empirical record, while the second term favors propositions that have received little credit mass; the latter is capped at $0.3$ to limit the influence of exploration. At the beginning of each epoch, the system computes $S$ for every proposition and injects up to $k=20$ of the highest-scoring ones into the solver prompt.

\subsection{Representative Knowledge-Base Entries}
\label{app:kb-examples}

Each entry below is drawn from the final knowledge-base snapshot and shown as a card in the style of Figure~\ref{fig:kb-samples}: notation is normalized, conditions and claims are preserved. The card header reports the entry's use count (total credit received, since each answer splits one unit of credit among the entries it cites), its accuracy $s/(s+f)$, its credibility $c=(s+1)/(s+f+2)$, and its final PUCT-style selection score $S$ from Appendix~\ref{app:puct-selection} evaluated with $N=400$ outcomes. Entries are grouped by topic and ordered by use count within each group.

\kbhead{kbaccopt!22}{Optimizer}

\begin{kbentry}{kbaccopt}{K0019}{uses 40.4 \,$\cdot$\, accuracy 81\% \,$\cdot$\, $c=0.797$ \,$\cdot$\, $S=0.911$}
Under fixed short budgets of roughly 500--2,000 steps and identical architectures, optimizer ranking is governed by the \kbob{effective learning-rate integral}, expressed as cumulative parameter displacement: Adam and AdamW produce sign-normalized updates with total displacement approximately \kbqt{$\eta T$}; Adagrad's decaying steps bound displacement at approximately \kbqt{$2\eta\sqrt{T}$}; and unaccelerated SGD on small MSE gradients yields only approximately \kbqt{$\sum_t\eta_t\lVert g_t\rVert$}.
\kbrk{The optimizer achieving the largest stable cumulative displacement tends to attain the greatest loss reduction.}
\end{kbentry}

\begin{kbentry}{kbaccopt}{K0016}{uses 16.4 \,$\cdot$\, accuracy 90\% \,$\cdot$\, $c=0.859$ \,$\cdot$\, $S=1.035$}
On continuous MSE regression targets at standard learning rates around $10^{-3}$, once the output head aligns near the target mean the \kbob{raw MSE gradients become small}, yet Adam's per-parameter gradient normalization still maintains \kbqt{effective updates of scale $\eta$}.
\kbrk{Under budgets of at most roughly 500 steps, Adam therefore reaches substantially lower test MSE than unaccelerated SGD at the same nominal rate.}
\end{kbentry}

\begin{kbentry}{kbaccopt}{K0156}{uses 13.1 \,$\cdot$\, accuracy 74\% \,$\cdot$\, $c=0.705$ \,$\cdot$\, $S=0.901$}
Comparing zero-initialized Adagrad with learning rate $\eta_g$ against sign-normalized Adam or AdamW with learning rate $\eta_a$, Adagrad produces larger initial updates of scale \kbqt{$\eta_g/\sqrt{t}$}, but Adam overtakes it in \kbob{cumulative displacement} after \kbqt{$t^*\approx(2\eta_g/\eta_a)^2$} steps.
\kbrk{On under-trained tasks with $T>t^*$, Adam therefore tends to achieve greater displacement and lower final loss.}
\end{kbentry}

\begin{kbentry}{kbaccopt}{K0012}{uses 5.8 \,$\cdot$\, accuracy 85\% \,$\cdot$\, $c=0.758$ \,$\cdot$\, $S=1.039$}
When fitting continuous targets with a \kbob{large constant offset $|\mathbb{E}[y]|$} relative to the initial output, SGD with momentum $m$ scales its effective learning rate to \kbqt{$\eta/(1-m)$}, with a bias-adaptation time constant of approximately \kbqt{$1/(2\eta/(1-m))$} steps.
\kbrk{Under tight budgets around $T=256$, this enables rapid early bias alignment and MSE reduction compared with unaccelerated optimizers.}
\end{kbentry}

\begin{kbentry}{kbaccopt}{K0069}{uses 2.4 \,$\cdot$\, accuracy 85\% \,$\cdot$\, $c=0.693$ \,$\cdot$\, $S=0.993$}
On continuous MSE regression targets, momentum SGD has effective learning rate \kbqt{$\eta/(1-m)$}, whereas the \kbob{per-step rate of zero-initialized Adagrad decays} approximately as \kbqt{$2\eta\sqrt{T}/T$}.
\kbrk{Under budgets of at most roughly 1,000 steps, when the former exceeds the latter by an order of magnitude, momentum SGD tends to achieve greater cumulative parameter displacement and loss reduction.}
\end{kbentry}

\kbhead{kbaccdat!22}{Dataset \& loss}

\begin{kbentry}{kbaccdat}{K0044}{uses 17.1 \,$\cdot$\, accuracy 86\% \,$\cdot$\, $c=0.822$ \,$\cdot$\, $S=0.995$}
On classification tasks with \kbob{zero linear baseline signal}, such as XOR, initial cross-entropy remains near the chance baseline \kbqt{$\ln 2\approx0.693$} until parameters move sufficiently far from initialization to form nonlinear feature interactions.
\kbrk{Under short budgets of at most roughly 256 steps, candidate progress is therefore ordered largely by stable effective step size.}
\end{kbentry}

\kbhead{kbaccarc!22}{Architecture}

\begin{kbentry}{kbaccarc}{K0038}{uses 7.1 \,$\cdot$\, accuracy 65\% \,$\cdot$\, $c=0.619$ \,$\cdot$\, $S=0.878$}
On complex two-dimensional classification tasks with \kbob{high-frequency decision boundaries}, such as multi-turn spirals, deep normalized residual MLPs with \kbqt{at least three residual blocks} achieve higher decision-boundary curvature and faster cross-entropy reduction per step than shallow MLPs of comparable width.
\kbrk{The latter require substantially greater width and longer budgets to resolve intricate boundary turns.}
\end{kbentry}

\begin{kbentry}{kbaccarc}{K0052}{uses 6.7 \,$\cdot$\, accuracy 63\% \,$\cdot$\, $c=0.601$ \,$\cdot$\, $S=0.867$}
For \kbob{first-order lookup tasks} such as bigram next-token prediction, a \kbqt{single}-layer transformer maps token embeddings directly to output logits, while a \kbqt{four}-layer post-LayerNorm stack adds optimization overhead and backpropagation attenuation.
\kbrk{Under budgets of at most roughly 256 steps at standard learning rates around $10^{-3}$, the shallow model reduces cross-entropy faster.}
\end{kbentry}

\begin{kbentry}{kbaccarc}{K0259}{uses 0.7 \,$\cdot$\, accuracy 100\% \,$\cdot$\, $c=0.623$ \,$\cdot$\, $S=0.923$}
On tabular classification with \kbob{smooth additive nonlinear target structure}, an intermediate-width shallow MLP (roughly \kbqt{width 64, two hidden layers}) supplies parallel coordinate-wise features, whereas a narrow deep MLP (roughly \kbqt{width 32, five or more hidden layers}) of comparable parameter count spreads the required optimization displacement across more layers.
\kbrk{Around 1,024 steps with AdamW near $3\times10^{-3}$, the shallow wide MLP can outperform the narrow deep one.}
\end{kbentry}

\begin{kbentry}{kbaccarc}{K0214}{uses 0.6 \,$\cdot$\, accuracy 100\% \,$\cdot$\, $c=0.615$ \,$\cdot$\, $S=0.915$}
For univariate targets with \kbob{several oscillations}, such as $x\cos(k\pi x)$ with \kbqt{$k\geq4$}, a single-hidden-layer network cannot readily generate high input frequencies because its first-layer weights must grow substantially from their default scale; residual stacks of \kbqt{depth at least four} with LayerNorm synthesize high-frequency curvature compositionally.
\kbrk{Trained for at most roughly 512 steps at adaptive learning rates around $3\times10^{-4}$, the deep residual stack fits such targets faster than shallow architectures of comparable size.}
\end{kbentry}

\subsection{Analytical Decision Tree}
\label{app:decision-tree-details}

This section visualizes the analytical decision policy introduced in Section~\ref{knowledge-base}. The flowchart summarizes an update-regime rule set; the underlying policy reaches 78.2\% dataset-grouped out-of-fold top-1 accuracy (391/500).

\begin{figure}[h]
\begin{center}
\includegraphics[width=0.82\linewidth]{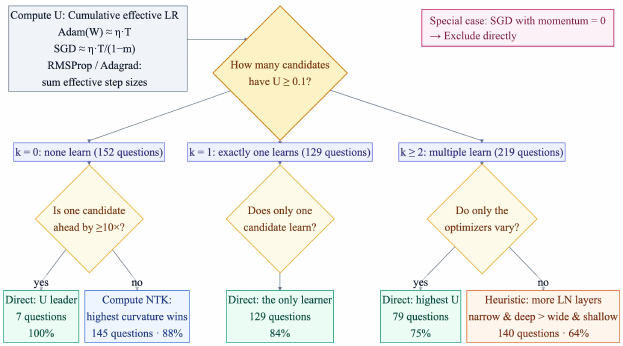}
\end{center}
\caption{Analytical decision tree summarizing an update-regime policy. The underlying development-stage policy reaches 78.2\% dataset-grouped out-of-fold top-1 accuracy (391/500). The simplified flowchart was not independently evaluated as a frozen predictor.}
\label{fig:decision-trees}
\end{figure}

\section{Full Benchmark Results}
\label{app:benchmark-results}

Table~\ref{tab:benchmark-results} reports the complete 500-question benchmark
evaluation results: overall accuracy and accuracy on each question type
(170 architecture-only, 168 optimizer-only and 162 mixed questions). Cell shading
uses one colour scale for all columns, from white at chance ($33.3\%$) to the
darkest shade at $90\%$, so darker cells indicate stronger performance and the
gap between the optimizer and architecture columns is visible at a glance. The
best-human row ($^\dagger$) is measured on the 50-question launch subset (22
architecture-only, 16 optimizer-only and 12 mixed questions); two participants
tie at the best score, and the per-type breakdown is taken from the one whose
item-level answers are fully recorded.

\begin{table}[h]
\caption{Complete 500-question benchmark results (\%). Shading runs from white at chance to dark at $90\%$; darker is better.}
\label{tab:benchmark-results}
\begin{center}
\setlength{\tabcolsep}{5pt}
\renewcommand{\arraystretch}{1.1}
\definecolor{aiqheat}{HTML}{2E5A85}
\begin{tabular}{lcccc}
\multicolumn{1}{c}{\bf Model} & \multicolumn{1}{c}{\bf Overall} & \multicolumn{1}{c}{\bf Architecture} & \multicolumn{1}{c}{\bf Optimizer} & \multicolumn{1}{c}{\bf Mixed} \\ \hline
GPT-5.6 Sol & \cellcolor{aiqheat!65}76.4 & \cellcolor{aiqheat!37}58 & \cellcolor{aiqheat!80}\textcolor{white}{87} & \cellcolor{aiqheat!77}\textcolor{white}{85} \\
Claude Opus 5 & \cellcolor{aiqheat!64}76.0 & \cellcolor{aiqheat!42}61 & \cellcolor{aiqheat!79}\textcolor{white}{86} & \cellcolor{aiqheat!71}\textcolor{white}{81} \\
Claude Fable 5 & \cellcolor{aiqheat!58}72.2 & \cellcolor{aiqheat!21}48 & \cellcolor{aiqheat!80}\textcolor{white}{87} & \cellcolor{aiqheat!74}\textcolor{white}{83} \\
DeepSeek V4 Flash & \cellcolor{aiqheat!56}70.6 & \cellcolor{aiqheat!21}48 & \cellcolor{aiqheat!78}\textcolor{white}{85} & \cellcolor{aiqheat!69}\textcolor{white}{80} \\
DeepSeek V4 Pro & \cellcolor{aiqheat!53}68.4 & \cellcolor{aiqheat!15}44 & \cellcolor{aiqheat!75}\textcolor{white}{83} & \cellcolor{aiqheat!69}\textcolor{white}{79} \\
GPT-5.6 Luna & \cellcolor{aiqheat!52}67.8 & \cellcolor{aiqheat!14}42 & \cellcolor{aiqheat!78}\textcolor{white}{85} & \cellcolor{aiqheat!65}77 \\
Gemini 3.1 Pro & \cellcolor{aiqheat!50}66.8 & \cellcolor{aiqheat!13}42 & \cellcolor{aiqheat!78}\textcolor{white}{85} & \cellcolor{aiqheat!61}74 \\
GPT-6 Astra & \cellcolor{aiqheat!49}66.2 & \cellcolor{aiqheat!7}38 & \cellcolor{aiqheat!76}\textcolor{white}{84} & \cellcolor{aiqheat!66}77 \\
GLM-5.2 & \cellcolor{aiqheat!49}66.0 & \cellcolor{aiqheat!9}39 & \cellcolor{aiqheat!77}\textcolor{white}{85} & \cellcolor{aiqheat!62}75 \\
Claude Sonnet 5 & \cellcolor{aiqheat!48}65.4 & \cellcolor{aiqheat!7}38 & \cellcolor{aiqheat!75}\textcolor{white}{83} & \cellcolor{aiqheat!63}75 \\
GPT-4o & \cellcolor{aiqheat!40}60.2 & \cellcolor{aiqheat!29}52 & \cellcolor{aiqheat!55}70 & \cellcolor{aiqheat!37}58 \\
DeepSeek R1 & \cellcolor{aiqheat!38}59.0 & \cellcolor{aiqheat!12}41 & \cellcolor{aiqheat!57}71 & \cellcolor{aiqheat!47}65 \\
Gemini 2.5 Pro & \cellcolor{aiqheat!34}55.8 & \cellcolor{aiqheat!6}37 & \cellcolor{aiqheat!54}70 & \cellcolor{aiqheat!42}61 \\
Llama 3.1 70B & \cellcolor{aiqheat!18}45.6 & \cellcolor{aiqheat!23}49 & \cellcolor{aiqheat!26}51 & \cellcolor{aiqheat!6}37 \\
MiniCPM5 1B & \cellcolor{aiqheat!8}39.0 & \cellcolor{aiqheat!12}41 & \cellcolor{aiqheat!18}45 & \cellcolor{aiqheat!0}30 \\
Qwen3.5 0.8B & \cellcolor{aiqheat!5}36.4 & \cellcolor{aiqheat!6}37 & \cellcolor{aiqheat!0}32 & \cellcolor{aiqheat!10}40 \\
\hline
Best human$^{\dagger}$ & \cellcolor{aiqheat!49}66.0 & \cellcolor{aiqheat!45}64 & \cellcolor{aiqheat!62}75 & \cellcolor{aiqheat!38}58 \\
Jev & \cellcolor{aiqheat!35}56.8 & \cellcolor{aiqheat!15}44 & \cellcolor{aiqheat!54}69 & \cellcolor{aiqheat!37}58 \\
Random choice & \cellcolor{aiqheat!0}33.3 & \cellcolor{aiqheat!0}33 & \cellcolor{aiqheat!0}33 & \cellcolor{aiqheat!0}33 \\
\end{tabular}

\end{center}
\end{table}

\section{Paired Label-Flipped Evaluation}
\label{app:label-flip}

For each pair, we independently sample two datasets from the same family with identical tensor shapes, together with a fresh set of three candidates. We execute all candidates for ten seeds on both datasets and retain a pair only when each side has a 10/10 winner with full seed-interval separation and the winning candidate differs between sides. The two sides are exchangeable: neither is designated as the original or counterfactual dataset.

This filter is deliberately strict. It ensures that the dataset change is sufficient to reverse the correct answer under the same criterion used by the reference panel. It also makes valid pairs rare. We obtain 52 pairs across univariate regression, multivariate regression, synthetic tabular classification, spiral classification, and bigram language modeling. XOR produces no accepted pair: under shape-matched sampling, its relevant interaction order and calibration remain fixed, while changing the signal-coordinate positions is largely symmetric for an MLP. Across 13,169 sampled pairs and a subsequent nine-candidate search, no XOR pair survives the strict two-sided flip criterion.

The single-side accuracies in Table~\ref{tab:paired-flip-results} exhibit a directional imbalance: the finite accepted sample happens to place the familiar candidate-level favorite on side A more often than side B, and models select the favorite at similar rates on both sides, inheriting this imbalance. The paired, combined accuracy and answer-transition statistics are the appropriate estimands.

\begin{table}[h]
\caption{Performance and answer consistency on paired, label-flipped questions}
\label{tab:paired-flip-results}
\begin{center}
\setlength{\tabcolsep}{3pt}
\renewcommand{\arraystretch}{1.15}
\begin{tabular}{@{}p{2.05in}cccc@{}}
\multicolumn{1}{c}{\bf Metric} &
\multicolumn{1}{c}{\shortstack{\bf Claude\\\bf Opus 5}} &
\multicolumn{1}{c}{\shortstack{\bf GPT-5.6\\\bf Sol}} &
\multicolumn{1}{c}{\shortstack{\bf GPT-5.6\\\bf Luna}} &
\multicolumn{1}{c}{\shortstack{\bf DeepSeek\\\bf V4 Flash}}
\\ \hline \\
Pairs
    & 52 & 52 & 52 & 52 \\
Side A accuracy
    & 0.538 & 0.519 & 0.635 & 0.615 \\
Side B accuracy
    & 0.423 & 0.404 & 0.365 & 0.327 \\
Combined accuracy
    & 0.481 & 0.462 & 0.500 & 0.471 \\
Answer-change rate
    & 0.231 & 0.058 & 0.096 & 0.059 \\
Previously correct answers that follow the flip
    & 0.179 & 0.037 & 0.091 & 0.031 \\
Correct-answer stickiness
    & 0.714 & 0.926 & 0.909 & 0.969 \\
\end{tabular}
\end{center}
\end{table}

\section{Data-Flip Evaluation}
\label{app:dataflip}

\paragraph{Questions.} A data-flip pair uses two datasets from the same family
(identical tensor shapes) and one shared set of three candidates, executed for
ten seeds on both. A pair is kept only if each side has a $10/10$ winner with
full seed separation and the winner differs between the sides. The prompts of
a pair differ only in the dataset section; candidates and letters are
identical. A solver that ignores the dataset is right on at most one side of
each pair, so its accuracy is at most $50\%$. \textbf{Hard500}: 250 pairs, 50
per family. \textbf{AIQ-Hard50}: 25 pairs (5 per family, seed 42), 50
questions (32 mixed, 10 architecture-only, 8 optimizer-only); chance
$33.3\%$. The other 225 pairs form the knowledge-base learning set and share
no dataset or candidate with Hard50.

\paragraph{Metrics.} Accuracy over valid answers; $J=(3a-1)/2$ (chance $=0$).
Per pair: \emph{same pick} (same candidate on both sides), \emph{both right},
\emph{exactly one}, \emph{neither}. Random-guess reference: same pick $1/3$,
both right $1/9$.

\begin{table}[t]
\caption{AIQ-Hard50 results (\%). $J = (3a-1)/2$ is the judgment score
(chance $= 0$). Arch./Opt./Mixed are per-type accuracies (10, 8 and 32
questions respectively). $^\dagger$36 of 50 answers valid (repeated
reasoning-budget exhaustion at max effort); $^\ddagger$47 of 50 valid.}
\label{tab:hard50-results}
\begin{center}
\setlength{\tabcolsep}{4.5pt}
\renewcommand{\arraystretch}{1.12}
\begin{tabular}{@{}lccccc@{}}
\multicolumn{1}{c}{\bf Model} &
\multicolumn{1}{c}{\bf Acc.} &
\multicolumn{1}{c}{\bf $J$} &
\multicolumn{1}{c}{\bf Arch.} &
\multicolumn{1}{c}{\bf Opt.} &
\multicolumn{1}{c}{\bf Mixed}
\\ \hline \\
GPT-6 Astra (max)
    & \textbf{52.0} & \textbf{0.280} & 60.0 & 50.0 & 50.0 \\
Claude Fable 5
    & 48.0 & 0.220 & 50.0 & 50.0 & 46.9 \\
Claude Opus 5
    & 48.0 & 0.220 & 50.0 & 50.0 & 46.9 \\
DeepSeek V4 Flash
    & 48.0 & 0.220 & 50.0 & 50.0 & 46.9 \\
DeepSeek V4 Pro
    & 48.0 & 0.220 & 50.0 & 50.0 & 46.9 \\
Gemini 3.1 Pro (high)
    & 48.0 & 0.220 & 60.0 & 37.5 & 46.9 \\
GLM-5.2 (max)$^\dagger$
    & 47.2 & 0.208 & 50.0 & 50.0 & 46.2 \\
Kimi K3$^\ddagger$
    & 46.8 & 0.202 & 50.0 & 37.5 & 48.4 \\
Claude Sonnet 5
    & 44.0 & 0.160 & 40.0 & 50.0 & 43.8 \\
GPT-5.6 Sol
    & 44.0 & 0.160 & 50.0 & 50.0 & 40.6 \\
GPT-5.6 Luna
    & 40.0 & 0.100 & 30.0 & 50.0 & 40.6 \\
GPT-4o
    & 40.0 & 0.100 & 60.0 & 50.0 & 31.2 \\
Llama 3.1 70B
    & 38.0 & 0.070 & 50.0 & 50.0 & 31.2 \\
\hline
CART decision tree
    & 48.0 & 0.220 & --- & --- & --- \\
\end{tabular}
\end{center}
\end{table}

\begin{figure}[t]
\begin{center}
\includegraphics[width=0.9\linewidth]{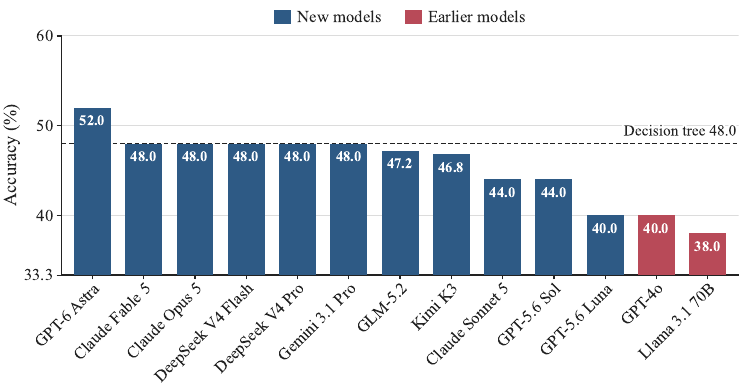}
\end{center}
\caption{AIQ-Hard50 accuracy. Dashed line: decision tree trained on the main
benchmark ($48.0\%$). Chance $33.3\%$; data-blind bound $50\%$.}
\label{fig:hard50-leaderboard}
\end{figure}

\begin{table}[t]
\caption{Per-pair outcomes on AIQ-Hard50 (25 flip pairs). \emph{Same pick}:
the model chose the same candidate on both sides despite the dataset change.
\emph{Both}/\emph{One}/\emph{Neither}: sides answered correctly. $^\dagger$15
and $^\ddagger$23 pairs with both sides answered (see
Table~\ref{tab:hard50-results}).}
\label{tab:hard50-pairs}
\begin{center}
\setlength{\tabcolsep}{5pt}
\renewcommand{\arraystretch}{1.12}
\begin{tabular}{@{}lcccc@{}}
\multicolumn{1}{c}{\bf Model} &
\multicolumn{1}{c}{\shortstack{\bf Same pick\\\bf (\%)}} &
\multicolumn{1}{c}{\shortstack{\bf Both\\\bf right (\%)}} &
\multicolumn{1}{c}{\shortstack{\bf Exactly\\\bf one (\%)}} &
\multicolumn{1}{c}{\shortstack{\bf Neither\\\bf (\%)}}
\\ \hline \\
GPT-6 Astra (max)
    & 84.0 & \textbf{12.0} & 80.0 & 8.0 \\
Claude Fable 5
    & 96.0 & 0.0 & 96.0 & 4.0 \\
Claude Opus 5
    & 100.0 & 0.0 & 96.0 & 4.0 \\
DeepSeek V4 Flash
    & 88.0 & 4.0 & 88.0 & 8.0 \\
DeepSeek V4 Pro
    & 84.0 & 8.0 & 80.0 & 12.0 \\
Gemini 3.1 Pro (high)
    & \textbf{76.0} & \textbf{12.0} & 72.0 & 16.0 \\
GLM-5.2 (max)$^\dagger$
    & 100.0 & 0.0 & 93.3 & 6.7 \\
Kimi K3$^\ddagger$
    & 95.7 & 0.0 & 91.3 & 8.7 \\
Claude Sonnet 5
    & 92.0 & 0.0 & 88.0 & 12.0 \\
GPT-5.6 Sol
    & 80.0 & 4.0 & 80.0 & 16.0 \\
GPT-5.6 Luna
    & \textbf{76.0} & 0.0 & 80.0 & 20.0 \\
GPT-4o
    & 80.0 & 4.0 & 72.0 & 24.0 \\
Llama 3.1 70B
    & 88.0 & 0.0 & 76.0 & 24.0 \\
\end{tabular}
\end{center}
\end{table}

\begin{table}[t]
\caption{Test-time scaling on AIQ-Hard50: accuracy (\%) by reasoning-effort
tier. $^\dagger$Sol does not support max; its top tier is xhigh.
$^\ddagger$Astra medium has 49 of 50 valid answers (one persistent transport
failure).}
\label{tab:hard50-tts}
\begin{center}
\setlength{\tabcolsep}{6pt}
\renewcommand{\arraystretch}{1.12}
\begin{tabular}{@{}lcccc@{}}
\multicolumn{1}{c}{\bf Model} &
\multicolumn{1}{c}{\bf Low} &
\multicolumn{1}{c}{\bf Medium} &
\multicolumn{1}{c}{\bf High} &
\multicolumn{1}{c}{\bf Max}
\\ \hline \\
GPT-6 Astra
    & 44.0 & 44.9$^\ddagger$ & 48.0 & \textbf{52.0} \\
Claude Opus 5
    & 46.0 & 48.0 & 48.0 & 48.0 \\
GPT-5.6 Luna
    & 42.0 & 44.0 & 40.0 & 48.0 \\
GPT-5.6 Sol$^\dagger$
    & 48.0 & 32.0 & 44.0 & 46.0 \\
\end{tabular}
\end{center}
\end{table}

\newcommand{\flipsub}[2]{%
  \begin{minipage}{\linewidth}\centering
  \includegraphics[width=0.9\linewidth]{figures/dataflip/ex_#1.pdf}\\[2pt]
  {\small #2}
  \end{minipage}}

\begin{figure}[p]
\centering
\flipsub{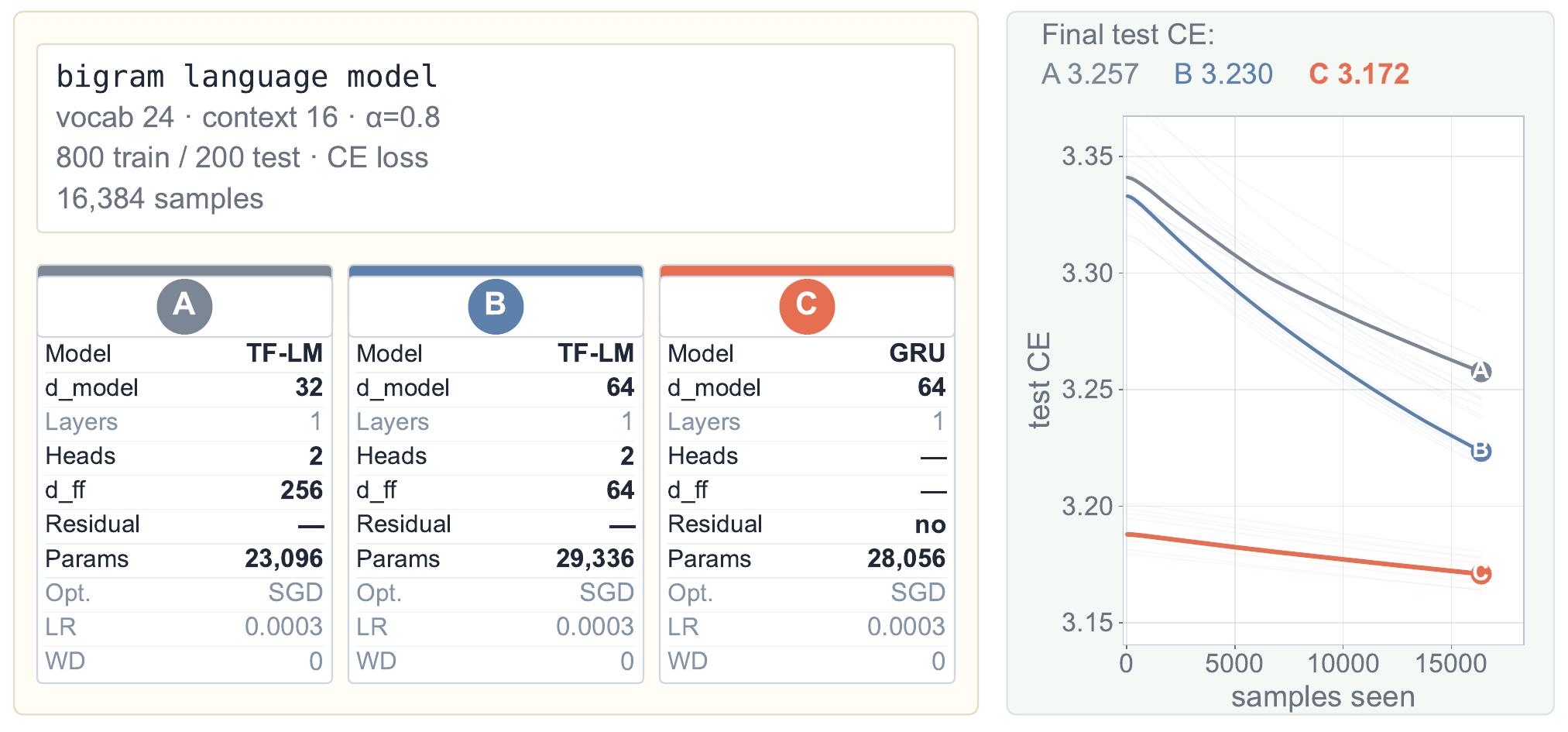}{(a) Bigram LM, $\alpha=0.8$. Architecture-only. Winner C.}\\[12pt]
\flipsub{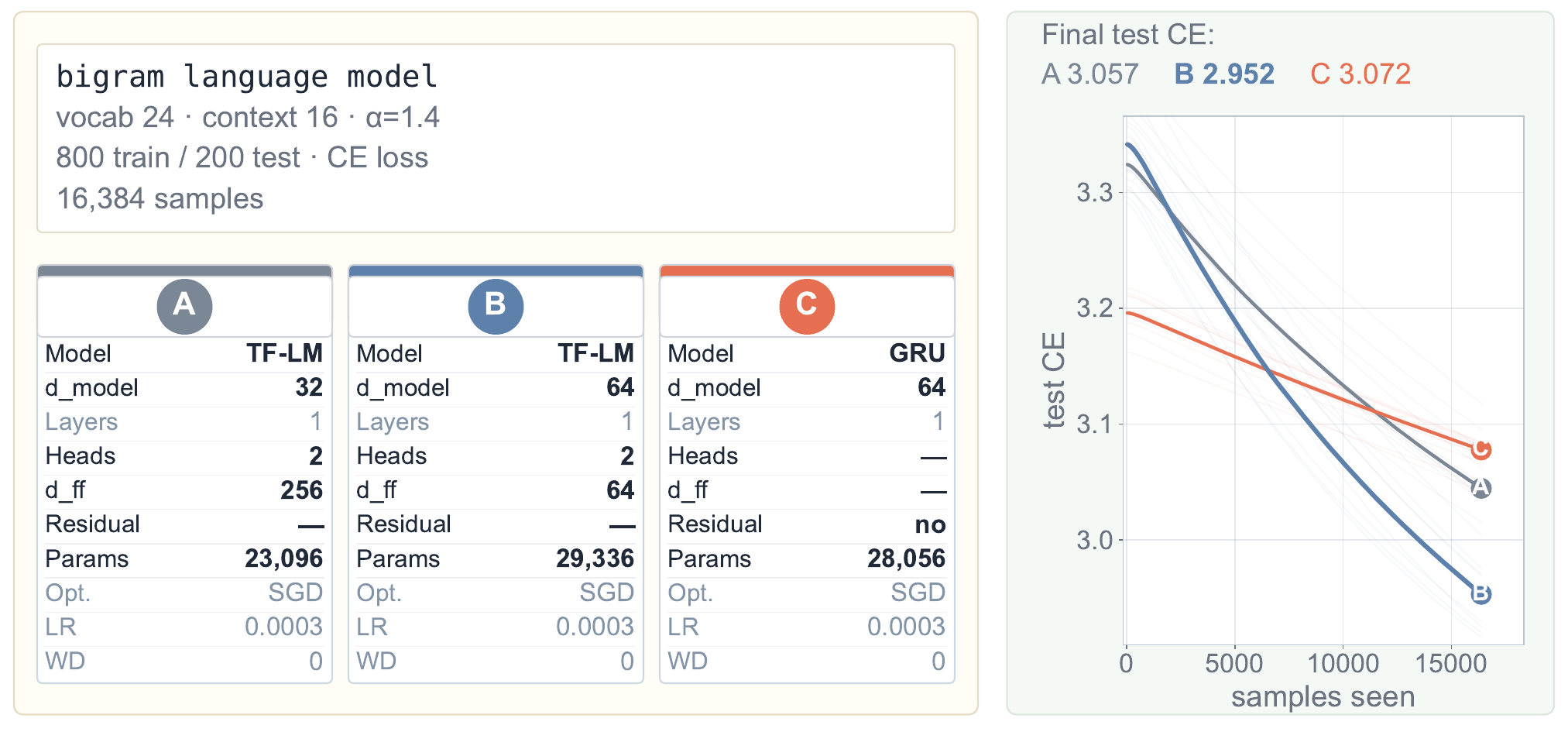}{(b) Same candidates as (a), $\alpha=1.4$. Winner B.}\\[12pt]
\flipsub{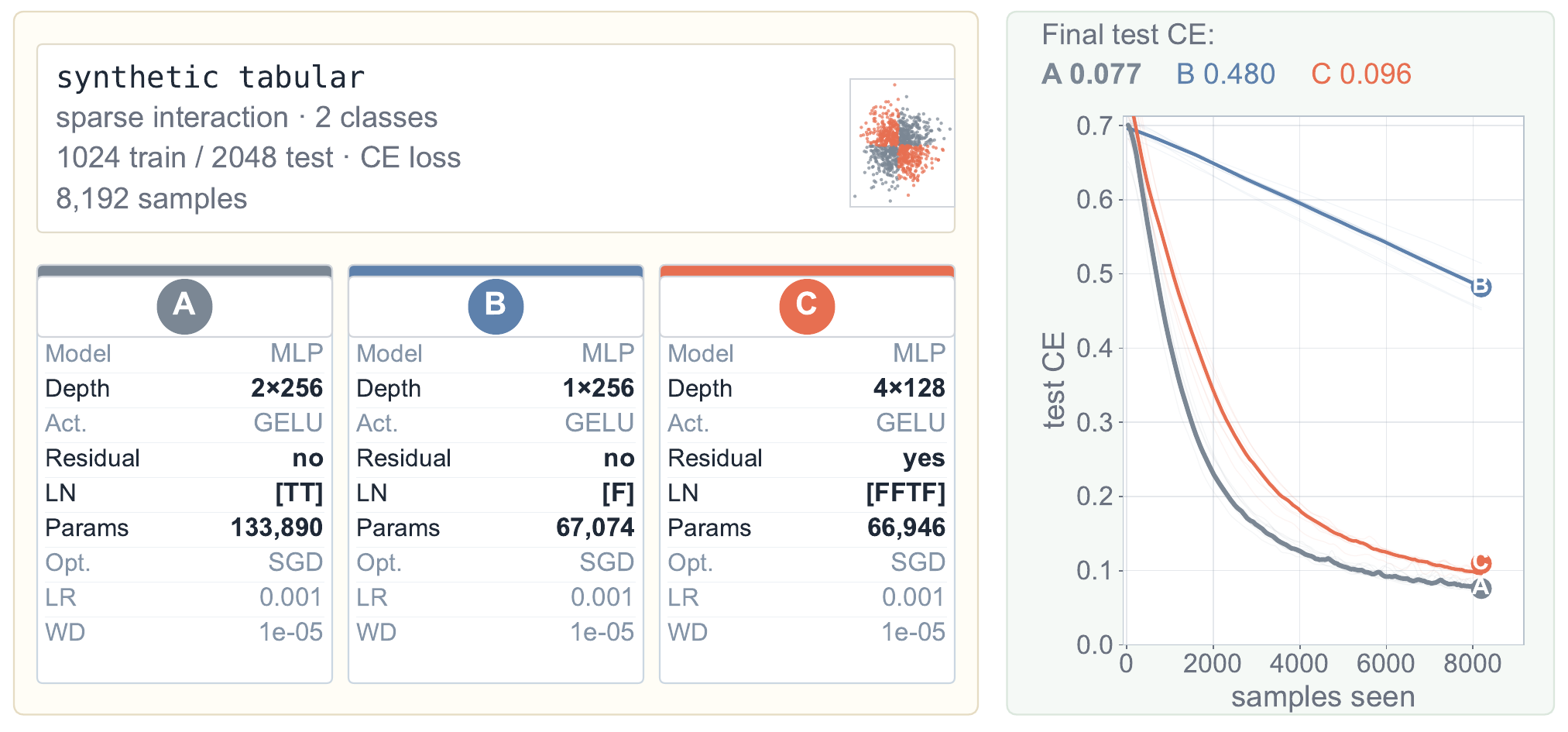}{(c) Tabular, sparse interaction. Architecture-only. Winner A.}
\caption{Example AIQ-Hard50 questions. Consecutive panels are the two sides of
one pair: (a,b), (c,d), (e,f), (g,h), (i,j). Left: dataset and candidates
(differing fields in bold). Right: executed ground truth over ten seeds
(median bold; winner bold in the legend).}
\label{fig:flip-examples}
\end{figure}

\begin{figure}[p]
\centering
\flipsub{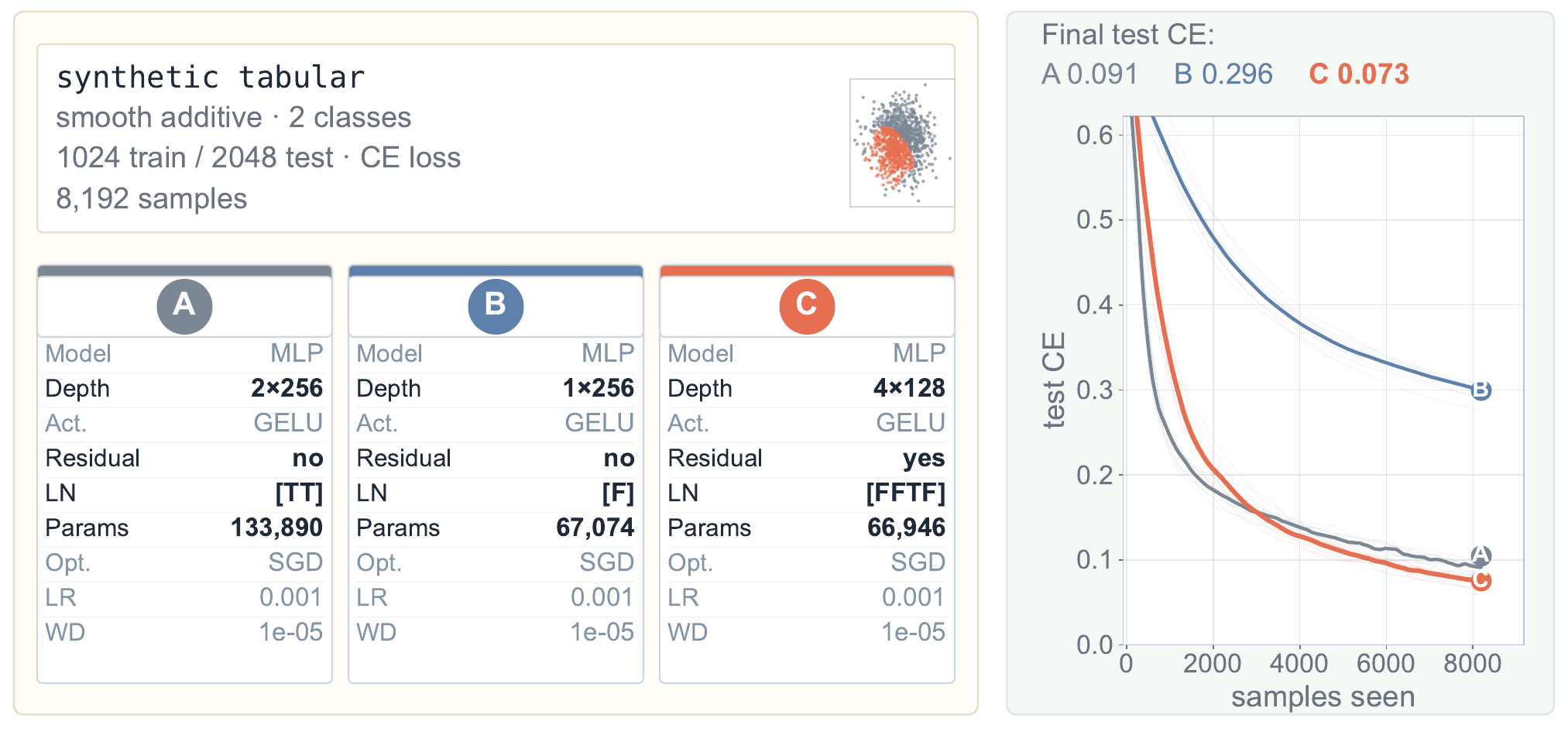}{(d) Same candidates as (c), smooth additive. Winner C.}\\[12pt]
\flipsub{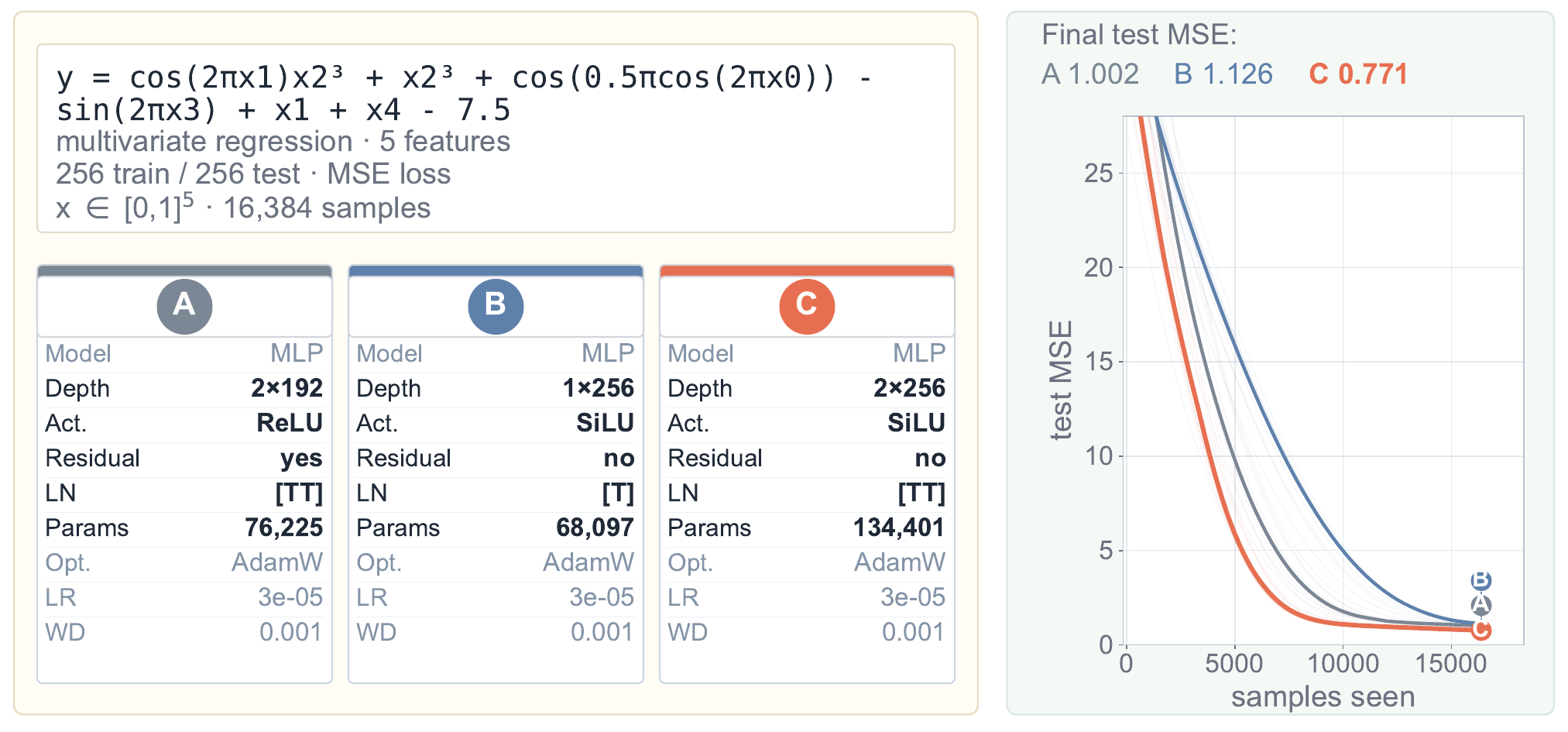}{(e) Multivariate regression. Architecture-only. Winner C.}\\[12pt]
\flipsub{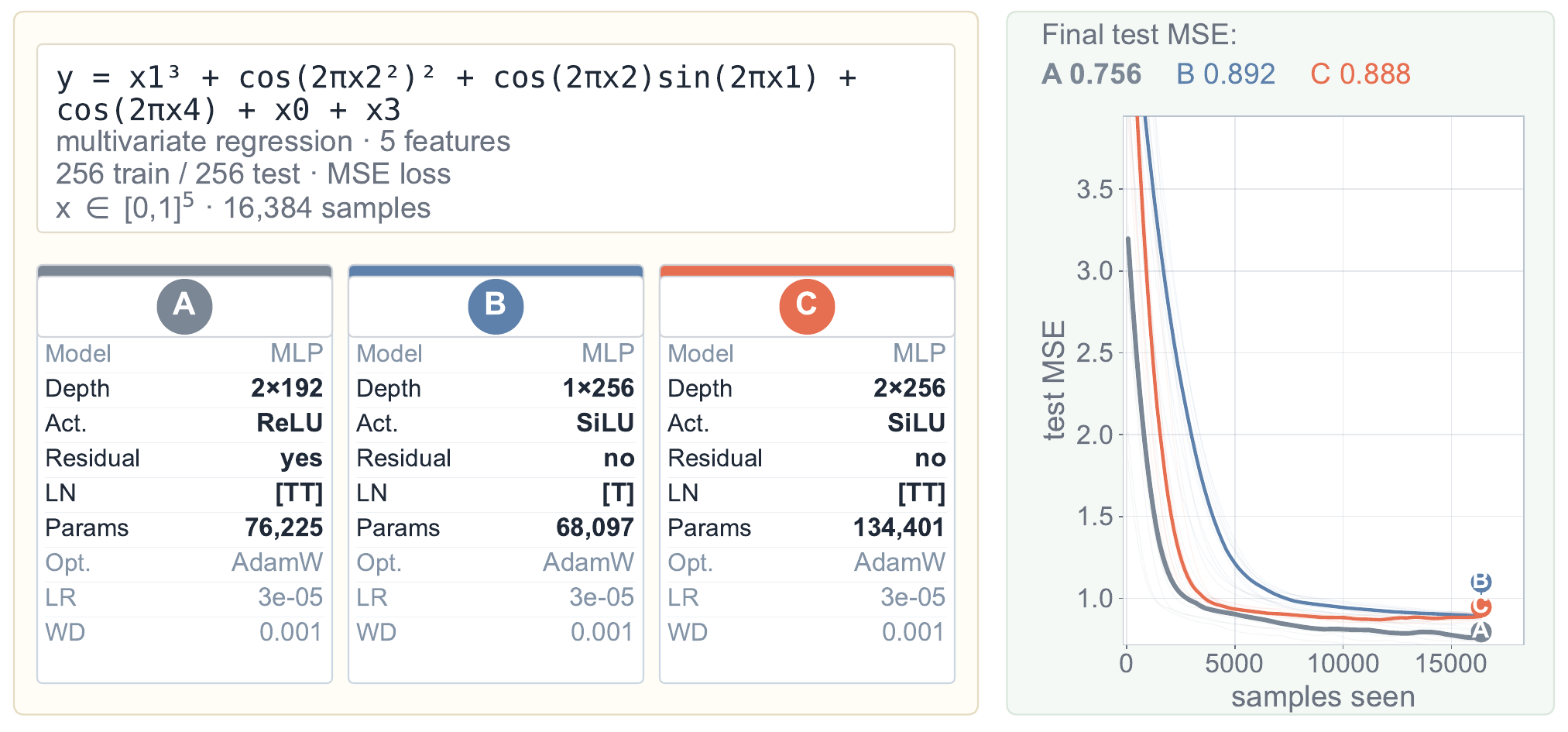}{(f) Same candidates as (e), different target. Winner A.}
\addtocounter{figure}{-1}
\caption{Example AIQ-Hard50 questions (continued).}
\end{figure}

\begin{figure}[p]
\centering
\flipsub{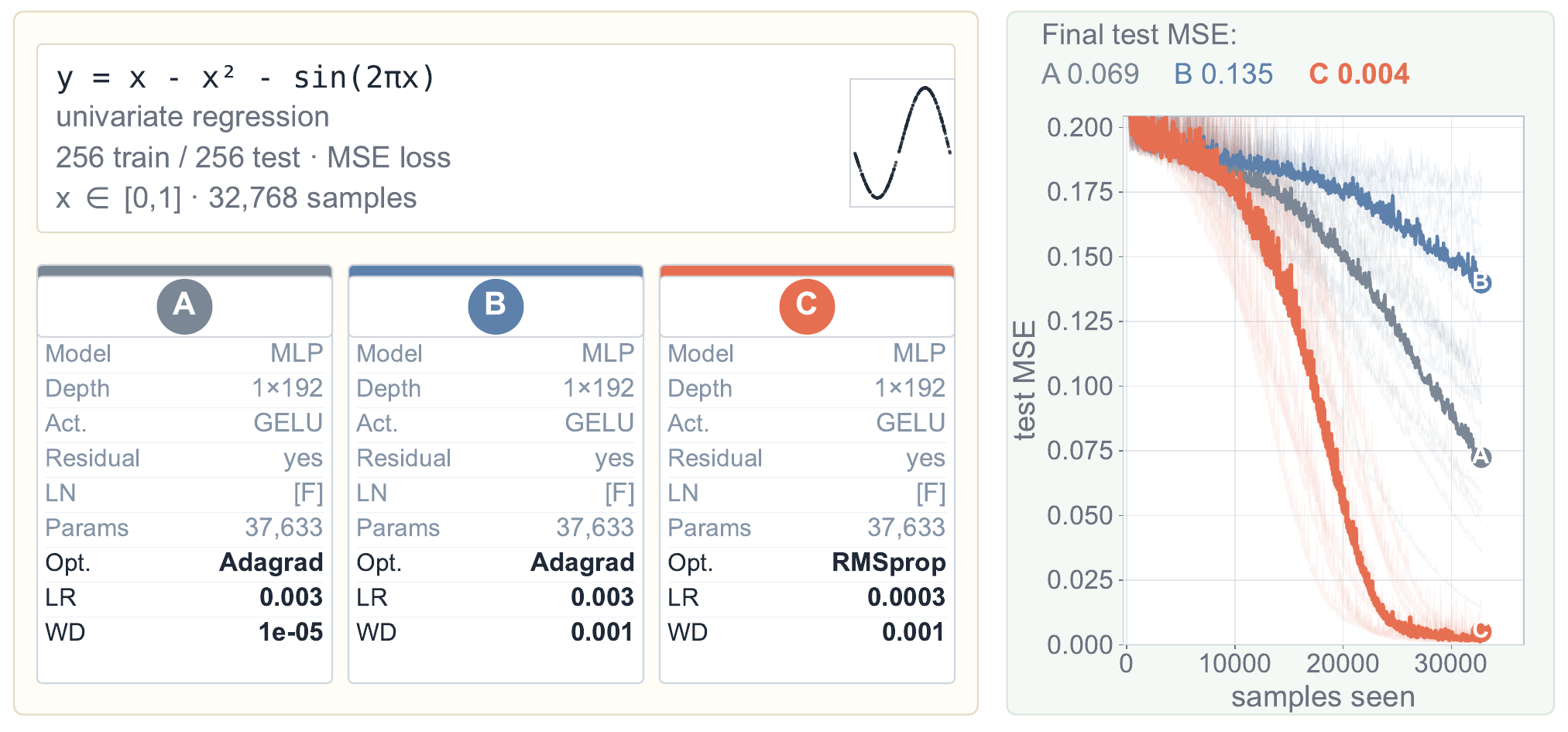}{(g) Univariate regression. Optimizer-only. Winner C.}\\[12pt]
\flipsub{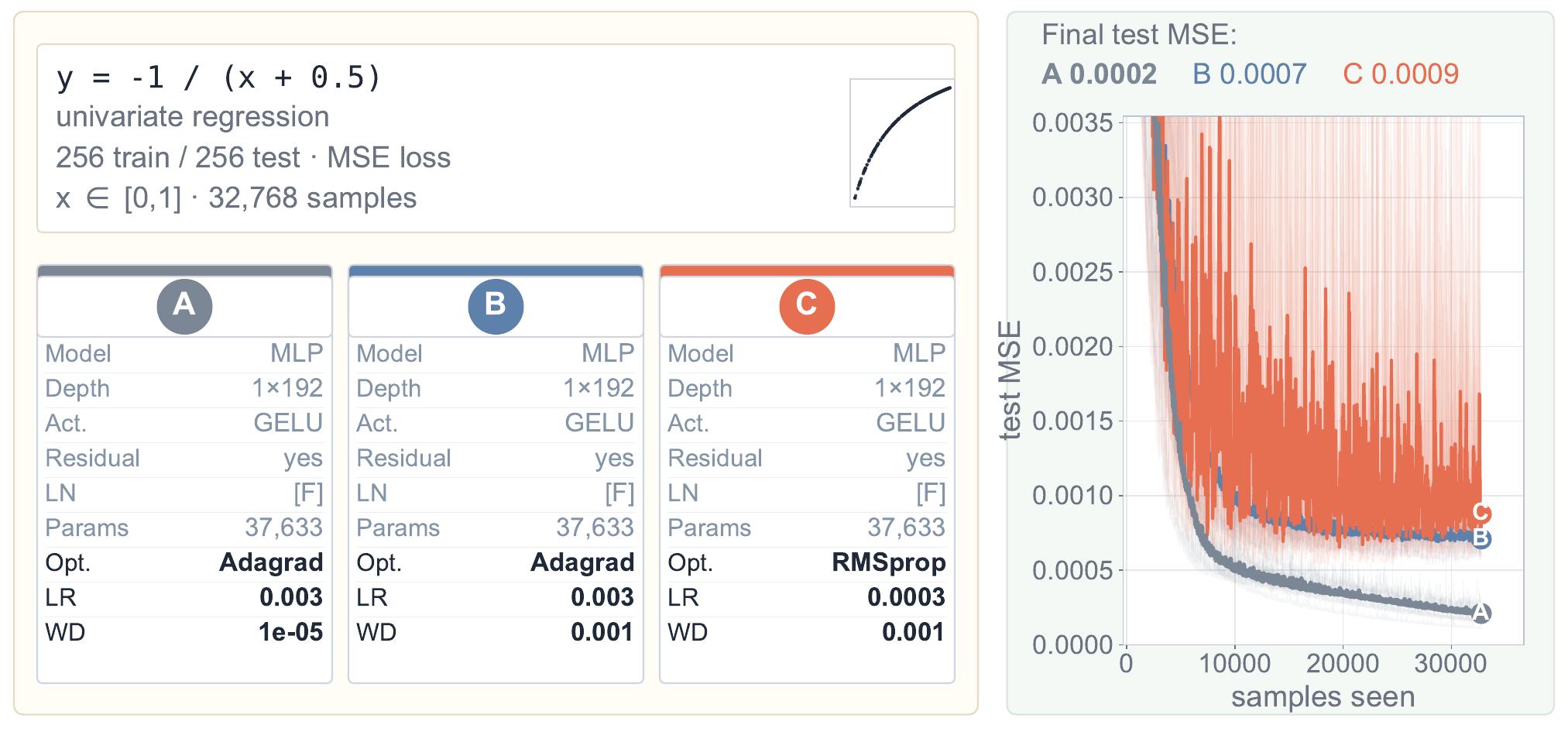}{(h) Same candidates as (g), different target. Winner A.}\\[12pt]
\flipsub{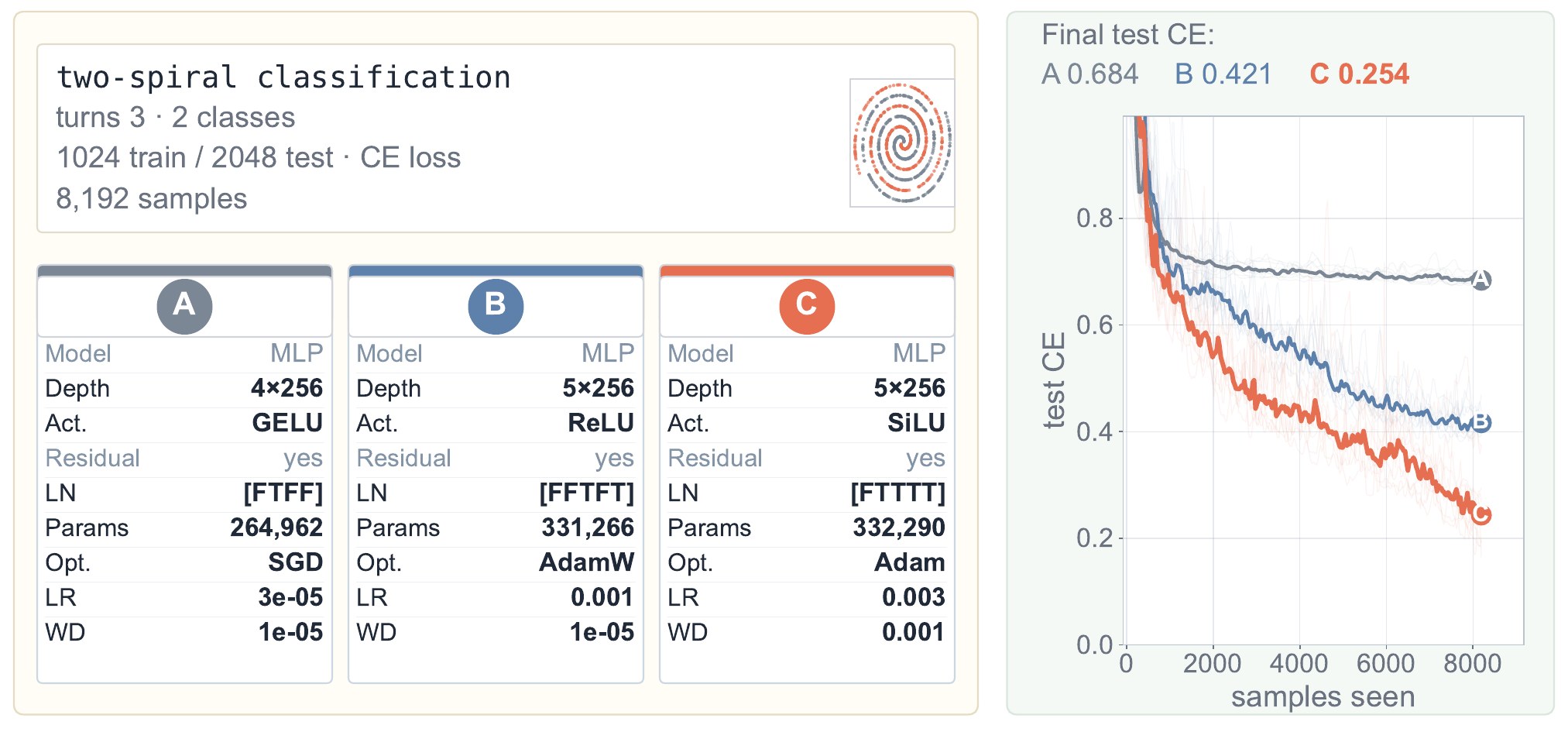}{(i) Two spirals, 3 turns. Mixed. Winner C.}
\addtocounter{figure}{-1}
\caption{Example AIQ-Hard50 questions (continued).}
\end{figure}

\begin{figure}[t]
\centering
\flipsub{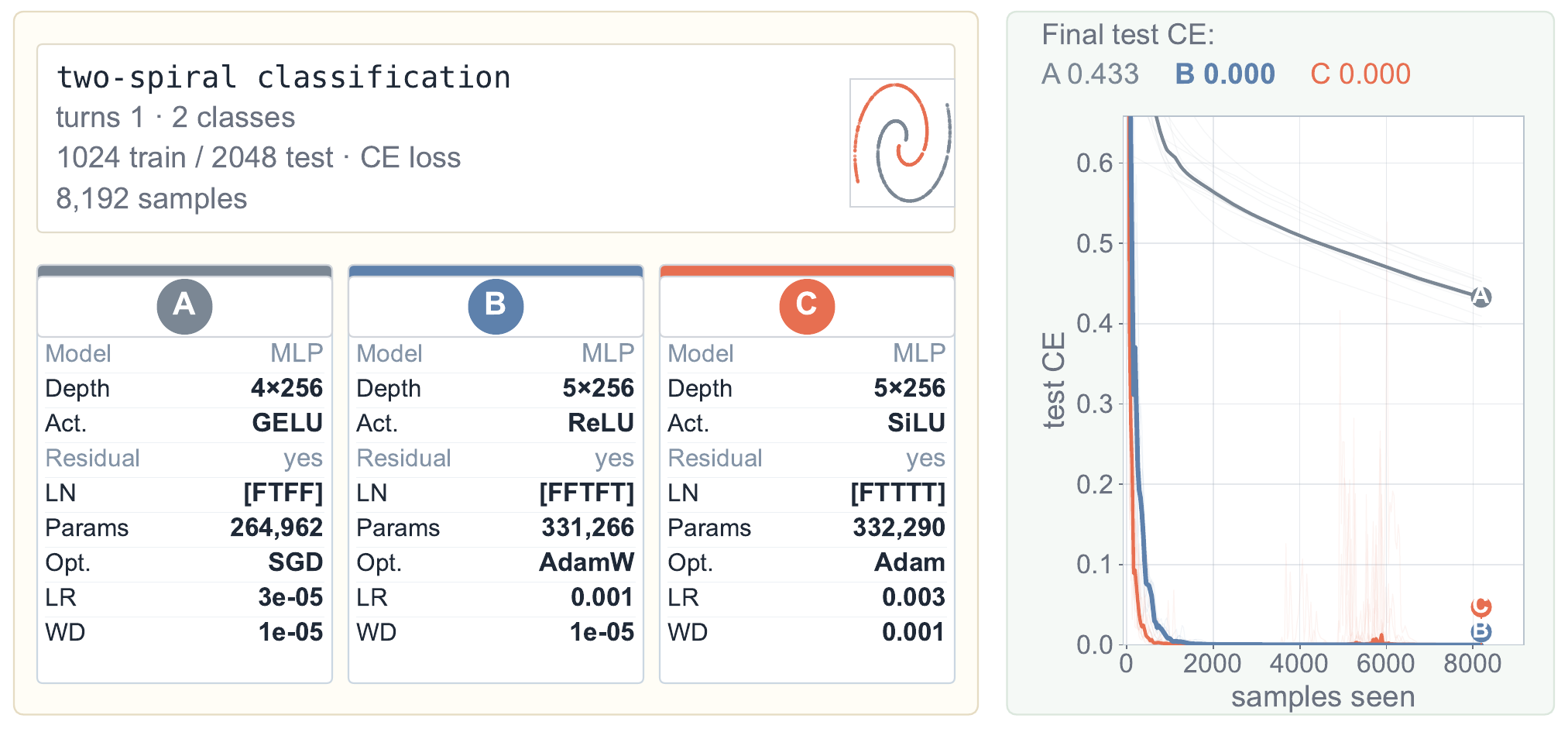}{(j) Same candidates as (i), 1 turn. Winner B.}
\addtocounter{figure}{-1}
\caption{Example AIQ-Hard50 questions (continued).}
\end{figure}

%% file: main.bbl
\begin{thebibliography}{31}
\providecommand{\natexlab}[1]{#1}
\providecommand{\url}[1]{\texttt{#1}}
\expandafter\ifx\csname urlstyle\endcsname\relax
  \providecommand{\doi}[1]{doi: #1}\else
  \providecommand{\doi}{doi: \begingroup \urlstyle{rm}\Url}\fi

\bibitem[Ben-David et~al.(2010)Ben-David, Blitzer, Crammer, Kulesza, Pereira,
  and Wortman~Vaughan]{bendavid2010theory}
Shai Ben-David, John Blitzer, Koby Crammer, Alex Kulesza, Fernando Pereira, and
  Jennifer Wortman~Vaughan.
\newblock A theory of learning from different domains.
\newblock \emph{Machine Learning}, 79\penalty0 (1--2):\penalty0 151--175, 2010.
\newblock \doi{10.1007/s10994-009-5152-4}.

\bibitem[Chan et~al.(2025)Chan, Chowdhury, Jaffe, Aung, Sherburn, Mays,
  Starace, Liu, Maksin, Patwardhan, Weng, and Madry]{chan2025mlebench}
Jun~Shern Chan, Neil Chowdhury, Oliver Jaffe, James Aung, Dane Sherburn, Evan
  Mays, Giulio Starace, Kevin Liu, Leon Maksin, Tejal Patwardhan, Lilian Weng,
  and Aleksander Madry.
\newblock {MLE-bench}: Evaluating machine learning agents on machine learning
  engineering.
\newblock In \emph{International Conference on Learning Representations}, 2025.

\bibitem[Domhan et~al.(2015)Domhan, Springenberg, and
  Hutter]{domhan2015speeding}
Tobias Domhan, Jost~Tobias Springenberg, and Frank Hutter.
\newblock Speeding up automatic hyperparameter optimization of deep neural
  networks by extrapolation of learning curves.
\newblock In \emph{Proceedings of the 24th International Joint Conference on
  Artificial Intelligence}, 2015.

\bibitem[Dong and Yang(2020)]{dong2020nasbench201}
Xuanyi Dong and Yi~Yang.
\newblock {NAS-Bench-201}: Extending the scope of reproducible neural
  architecture search.
\newblock In \emph{International Conference on Learning Representations}, 2020.

\bibitem[Falkner et~al.(2018)Falkner, Klein, and Hutter]{falkner2018bohb}
Stefan Falkner, Aaron Klein, and Frank Hutter.
\newblock {BOHB}: Robust and efficient hyperparameter optimization at scale.
\newblock In \emph{Proceedings of the 35th International Conference on Machine
  Learning}, pages 1437--1446, 2018.

\bibitem[Gebru et~al.(2021)Gebru, Morgenstern, Vecchione, Wortman~Vaughan,
  Wallach, Daum{\'e}~III, and Crawford]{gebru2021datasheets}
Timnit Gebru, Jamie Morgenstern, Briana Vecchione, Jennifer Wortman~Vaughan,
  Hanna Wallach, Hal Daum{\'e}~III, and Kate Crawford.
\newblock Datasheets for datasets.
\newblock \emph{Communications of the ACM}, 64\penalty0 (12):\penalty0 86--92,
  2021.
\newblock \doi{10.1145/3458723}.

\bibitem[Huang et~al.(2024)Huang, Vora, Liang, and
  Leskovec]{huang2024mlagentbench}
Qian Huang, Jian Vora, Percy Liang, and Jure Leskovec.
\newblock {MLAgentBench}: Evaluating language agents on machine learning
  experimentation.
\newblock In \emph{Proceedings of the 41st International Conference on Machine
  Learning}, 2024.

\bibitem[Jimenez et~al.(2024)Jimenez, Yang, Wettig, Yao, Pei, Press, and
  Narasimhan]{jimenez2024swebench}
Carlos~E. Jimenez, John Yang, Alexander Wettig, Shunyu Yao, Kexin Pei, Ofir
  Press, and Karthik Narasimhan.
\newblock {SWE-bench}: Can language models resolve real-world {GitHub} issues?
\newblock In \emph{International Conference on Learning Representations}, 2024.

\bibitem[Klein et~al.(2017)Klein, Falkner, Springenberg, and
  Hutter]{klein2017learningcurve}
Aaron Klein, Stefan Falkner, Jost~Tobias Springenberg, and Frank Hutter.
\newblock Learning curve prediction with bayesian neural networks.
\newblock In \emph{International Conference on Learning Representations}, 2017.

\bibitem[Koh et~al.(2021)Koh, Sagawa, Marklund, Xie, Zhang, Balsubramani, Hu,
  Yasunaga, Phillips, Gao, et~al.]{koh2021wilds}
Pang~Wei Koh, Shiori Sagawa, Henrik Marklund, Sang~Michael Xie, Marvin Zhang,
  Akshay Balsubramani, Weihua Hu, Michihiro Yasunaga, Richard~L. Phillips,
  Irena Gao, et~al.
\newblock {WILDS}: A benchmark of in-the-wild distribution shifts.
\newblock In \emph{Proceedings of the 38th International Conference on Machine
  Learning}, 2021.

\bibitem[Li et~al.(2018)Li, Jamieson, DeSalvo, Rostamizadeh, and
  Talwalkar]{li2018hyperband}
Lisha Li, Kevin Jamieson, Giulia DeSalvo, Afshin Rostamizadeh, and Ameet
  Talwalkar.
\newblock Hyperband: A novel bandit-based approach to hyperparameter
  optimization.
\newblock \emph{Journal of Machine Learning Research}, 18\penalty0
  (185):\penalty0 1--52, 2018.

\bibitem[Liu et~al.(2024)Liu, Yu, Zhang, Xu, Lei, Lai, Gu, Ding, Men, Yang,
  et~al.]{liu2024agentbench}
Xiao Liu, Hao Yu, Hanchen Zhang, Yifan Xu, Xuanyu Lei, Hanyu Lai, Yu~Gu,
  Hangliang Ding, Kaiwen Men, Kejuan Yang, et~al.
\newblock {AgentBench}: Evaluating {LLM}s as agents.
\newblock In \emph{International Conference on Learning Representations}, 2024.

\bibitem[Lu et~al.(2024)Lu, Lu, Lange, Foerster, Clune, and
  Ha]{lu2024aiscientist}
Chris Lu, Cong Lu, Robert~Tjarko Lange, Jakob Foerster, Jeff Clune, and David
  Ha.
\newblock The {AI} scientist: Towards fully automated open-ended scientific
  discovery.
\newblock \emph{arXiv preprint arXiv:2408.06292}, 2024.

\bibitem[Madaan et~al.(2023)Madaan, Tandon, Gupta, Hallinan, Gao, Wiegreffe,
  Alon, Dziri, Prabhumoye, Yang, Gupta, Majumder, Hermann, Welleck,
  Yazdanbakhsh, and Clark]{madaan2023selfrefine}
Aman Madaan, Niket Tandon, Prakhar Gupta, Skyler Hallinan, Luyu Gao, Sarah
  Wiegreffe, Uri Alon, Nouha Dziri, Shrimai Prabhumoye, Yiming Yang, Shashank
  Gupta, Bodhisattwa~Prasad Majumder, Katherine Hermann, Sean Welleck, Amir
  Yazdanbakhsh, and Peter Clark.
\newblock Self-refine: Iterative refinement with self-feedback.
\newblock In \emph{Advances in Neural Information Processing Systems}, 2023.

\bibitem[Paullada et~al.(2021)Paullada, Raji, Bender, Denton, and
  Hanna]{paullada2021data}
Amandalynne Paullada, Inioluwa~Deborah Raji, Emily~M. Bender, Emily Denton, and
  Alex Hanna.
\newblock Data and its (dis)contents: A survey of dataset development and use
  in machine learning research.
\newblock \emph{Patterns}, 2\penalty0 (11):\penalty0 100336, 2021.
\newblock \doi{10.1016/j.patter.2021.100336}.

\bibitem[Sambasivan et~al.(2021)Sambasivan, Kapania, Highfill, Akrong,
  Paritosh, and Aroyo]{sambasivan2021data}
Nithya Sambasivan, Shivani Kapania, Hannah Highfill, Diana Akrong, Praveen~K.
  Paritosh, and Lora~M. Aroyo.
\newblock {``Everyone Wants to Do the Model Work, Not the Data Work''}: Data
  cascades in high-stakes {AI}.
\newblock In \emph{Proceedings of the 2021 CHI Conference on Human Factors in
  Computing Systems}, pages 1--15, 2021.
\newblock \doi{10.1145/3411764.3445518}.

\bibitem[Shinn et~al.(2023)Shinn, Cassano, Gopinath, Narasimhan, and
  Yao]{shinn2023reflexion}
Noah Shinn, Federico Cassano, Ashwin Gopinath, Karthik Narasimhan, and Shunyu
  Yao.
\newblock Reflexion: Language agents with verbal reinforcement learning.
\newblock In \emph{Advances in Neural Information Processing Systems}, 2023.

\bibitem[Silver et~al.(2018)Silver, Hubert, Schrittwieser, Antonoglou, Lai,
  Guez, Lanctot, Sifre, Kumaran, Graepel, Lillicrap, Simonyan, and
  Hassabis]{silver2018alphazero}
David Silver, Thomas Hubert, Julian Schrittwieser, Ioannis Antonoglou, Matthew
  Lai, Arthur Guez, Marc Lanctot, Laurent Sifre, Dharshan Kumaran, Thore
  Graepel, Timothy Lillicrap, Karen Simonyan, and Demis Hassabis.
\newblock A general reinforcement learning algorithm that masters chess, shogi,
  and go through self-play.
\newblock \emph{Science}, 362\penalty0 (6419):\penalty0 1140--1144, 2018.
\newblock \doi{10.1126/science.aar6404}.

\bibitem[Snoek et~al.(2012)Snoek, Larochelle, and Adams]{snoek2012bayesian}
Jasper Snoek, Hugo Larochelle, and Ryan~P. Adams.
\newblock Practical bayesian optimization of machine learning algorithms.
\newblock In \emph{Advances in Neural Information Processing Systems}, 2012.

\bibitem[Starace et~al.(2025)Starace, Jaffe, Sherburn, Aung, Chan, Maksin,
  Dias, Mays, Kinsella, Thompson, Heidecke, Glaese, and
  Patwardhan]{starace2025paperbench}
Giulio Starace, Oliver Jaffe, Dane Sherburn, James Aung, Jun~Shern Chan, Leon
  Maksin, Rachel Dias, Evan Mays, Benjamin Kinsella, Wyatt Thompson, Johannes
  Heidecke, Amelia Glaese, and Tejal Patwardhan.
\newblock {PaperBench}: Evaluating {AI}'s ability to replicate {AI} research.
\newblock \emph{arXiv preprint arXiv:2504.01848}, 2025.

\bibitem[Wang et~al.(2023{\natexlab{a}})Wang, Xie, Jiang, Mandlekar, Xiao, Zhu,
  Fan, and Anandkumar]{wang2023voyager}
Guanzhi Wang, Yuqi Xie, Yunfan Jiang, Ajay Mandlekar, Chaowei Xiao, Yuke Zhu,
  Linxi Fan, and Anima Anandkumar.
\newblock Voyager: An open-ended embodied agent with large language models.
\newblock \emph{Transactions on Machine Learning Research}, 2023{\natexlab{a}}.

\bibitem[Wang et~al.(2023{\natexlab{b}})Wang, Wei, Schuurmans, Le, Chi, Narang,
  Chowdhery, and Zhou]{wang2023selfconsistency}
Xuezhi Wang, Jason Wei, Dale Schuurmans, Quoc~V. Le, Ed~H. Chi, Sharan Narang,
  Aakanksha Chowdhery, and Denny Zhou.
\newblock Self-consistency improves chain of thought reasoning in language
  models.
\newblock In \emph{International Conference on Learning Representations},
  2023{\natexlab{b}}.

\bibitem[Wei et~al.(2022)Wei, Wang, Schuurmans, Bosma, Xia, Chi, Le, and
  Zhou]{wei2022chainofthought}
Jason Wei, Xuezhi Wang, Dale Schuurmans, Maarten Bosma, Fei Xia, Ed~Chi, Quoc
  Le, and Denny Zhou.
\newblock Chain-of-thought prompting elicits reasoning in large language
  models.
\newblock In \emph{Advances in Neural Information Processing Systems}, 2022.

\bibitem[Whang et~al.(2023)Whang, Roh, Song, and Lee]{whang2023data}
Steven~Euijong Whang, Yuji Roh, Hwanjun Song, and Jae-Gil Lee.
\newblock Data collection and quality challenges in deep learning: A
  data-centric {AI} perspective.
\newblock \emph{The VLDB Journal}, 32\penalty0 (4):\penalty0 791--813, 2023.
\newblock \doi{10.1007/s00778-022-00775-9}.

\bibitem[White et~al.(2021)White, Zela, Ru, Liu, and
  Hutter]{white2021predictors}
Colin White, Arber Zela, Binxin Ru, Yang Liu, and Frank Hutter.
\newblock How powerful are performance predictors in neural architecture
  search?
\newblock In \emph{Advances in Neural Information Processing Systems}, 2021.

\bibitem[Yao et~al.(2023{\natexlab{a}})Yao, Yu, Zhao, Shafran, Griffiths, Cao,
  and Narasimhan]{yao2023tree}
Shunyu Yao, Dian Yu, Jeffrey Zhao, Izhak Shafran, Thomas~L. Griffiths, Yuan
  Cao, and Karthik Narasimhan.
\newblock Tree of thoughts: Deliberate problem solving with large language
  models.
\newblock In \emph{Advances in Neural Information Processing Systems},
  2023{\natexlab{a}}.

\bibitem[Yao et~al.(2023{\natexlab{b}})Yao, Zhao, Yu, Du, Shafran, Narasimhan,
  and Cao]{yao2023react}
Shunyu Yao, Jeffrey Zhao, Dian Yu, Nan Du, Izhak Shafran, Karthik Narasimhan,
  and Yuan Cao.
\newblock {ReAct}: Synergizing reasoning and acting in language models.
\newblock In \emph{International Conference on Learning Representations},
  2023{\natexlab{b}}.

\bibitem[Ying et~al.(2019)Ying, Klein, Christiansen, Real, Murphy, and
  Hutter]{ying2019nasbench101}
Chris Ying, Aaron Klein, Eric Christiansen, Esteban Real, Kevin Murphy, and
  Frank Hutter.
\newblock {NAS-Bench-101}: Towards reproducible neural architecture search.
\newblock In \emph{Proceedings of the 36th International Conference on Machine
  Learning}, 2019.

\bibitem[Zela et~al.(2022)Zela, Siems, Zimmer, Lukasik, Keuper, and
  Hutter]{zela2022surrogate}
Arber Zela, Julien Siems, Lucas Zimmer, Jovita Lukasik, Margret Keuper, and
  Frank Hutter.
\newblock Surrogate nas benchmarks: Going beyond the limited search spaces of
  tabular nas benchmarks.
\newblock In \emph{International Conference on Learning Representations}, 2022.

\bibitem[Zelikman et~al.(2022)Zelikman, Wu, Mu, and Goodman]{zelikman2022star}
Eric Zelikman, Yuhuai Wu, Jesse Mu, and Noah~D. Goodman.
\newblock {STaR}: Bootstrapping reasoning with reasoning.
\newblock In \emph{Advances in Neural Information Processing Systems}, 2022.

\bibitem[Zhao et~al.(2024)Zhao, Huang, Xu, Lin, Liu, and Huang]{zhao2024expel}
Andrew Zhao, Daniel Huang, Quentin Xu, Matthieu Lin, Yong-Jin Liu, and Gao
  Huang.
\newblock {ExpeL}: {LLM} agents are experiential learners.
\newblock In \emph{Proceedings of the AAAI Conference on Artificial
  Intelligence}, volume~38, 2024.
\newblock \doi{10.1609/aaai.v38i17.29936}.

\end{thebibliography}
